\documentclass{article}

\usepackage[final,nonatbib]{nips_2018}

\usepackage[utf8]{inputenc} 
\usepackage[T1]{fontenc}    
\usepackage{hyperref}       
\usepackage{url}            
\usepackage{booktabs}       
\usepackage{amsfonts}       
\usepackage{nicefrac}       
\usepackage{microtype}      
\usepackage{amssymb}

\usepackage{lipsum}
\usepackage{xspace}
\usepackage{color}
\usepackage{tikz}
\usepackage{amsmath}
\usepackage{bm}
\usepackage{array}
\newcolumntype{C}[1]{>{\centering\arraybackslash}m{#1}}

\definecolor{mygreen}{rgb}{0.2, 0.7, 0.2}
\definecolor{myorange}{rgb}{0.9, 0.5, 0.0}
\definecolor{mypurple}{rgb}{0.5, 0, 0.5}

\newcommand{\nbasis}{m} 
\newcommand{\nconv}{d} 
\newcommand{\nclass}{Q} 
\newcommand{\nobs}{n} 

\newcommand{\LL}{\mathcal{L}}
\newcommand{\E}{\mathrm{E}}

\newcommand{\diag}{\mathrm{diag}}

\newcommand{\norm}{\mathcal{N}}

\newcommand{\cvect}{\mathbf{c}}

\newcommand{\gvect}{\mathbf{g}}

\newcommand{\zvect}{\mathbf{z}}
\newcommand{\xvect}{\mathbf{x}}
\newcommand{\yvect}{\mathbf{y}}

\newcommand{\zerovect}{\mathbf{0}}

\newcommand{\Dmat}{\mathbf{D}}
\newcommand{\Xmat}{\mathbf{X}}

\newcommand{\Ymat}{\mathbf{Y}}
\newcommand{\Fmat}{\mathbf{F}}

\newcommand{\Wmat}{\mathbf{W}}
\newcommand{\Gmat}{\mathbf{G}}
\newcommand{\Hmat}{\mathbf{H}}
\newcommand{\Cmat}{\mathbf{C}}

\newcommand{\Imat}{\mathbf{I}}

\newcommand{\Mmat}{\mathbf{M}}
\newcommand{\Phimat}{\boldsymbol{\Phi}}

\newcommand{\Omegamat}{\boldsymbol{\Omega}}
\newcommand{\Psimat}{\boldsymbol{\Psi}}

\newcommand{\Lambdamat}{\boldsymbol{\Lambda}}

\newcommand{\thetavect}{\boldsymbol{\theta}}

\newcommand{\omegavect}{\boldsymbol{\omega}}

\newcommand{\varepsilonvect}{\boldsymbol{\epsilon}}

\newcommand{\bigO}{\mathcal{O}}

\newcommand{\name}[1]{{\textsc{#1}}\xspace}

\newcommand{\mnist}{\name{mnist}} 
\newcommand{\notmnist}{\name{not-mnist}} 
\newcommand{\cifart}{\name{cifar10}}
\newcommand{\cifarh}{\name{cifar100}}

\newcommand{\arccosine}{\name{arc-cosine}}

\newcommand{\gp}{\name{gp}}
\newcommand{\gps}{\textsc{gp}s\xspace}
\newcommand{\dgp}{\name{dgp}}
\newcommand{\dgps}{\textsc{dgp}s\xspace}
\newcommand{\dnn}{\name{dnn}}
\newcommand{\dnns}{\textsc{dnn}s\xspace}

\newcommand{\convweights}{\Psimat}

\newcommand{\relu}{{\textsc{r}}e\name{lu}}

\newcommand{\rbf}{\name{rbf}}

\newcommand{\mcd}{\name{mcd}}

\newcommand{\mnll}{\name{mnll}}

\newcommand{\err}{\name{err}}
\newcommand{\brier}{\name{brier}}
\newcommand{\ece}{\name{ece}}

\newcommand{\lenet}{\name{LeNet}}
\newcommand{\resnet}{\name{resnet}}

\newcommand{\cnn}{\name{cnn}}
\newcommand{\cnns}{\textsc{cnn}s\xspace}
\newcommand{\gpdnn}{\name{gpdnn}}
\newcommand{\cgp}{\name{cgp}}
\newcommand{\svdkl}{\name{svdkl}}

\newcommand{\sorf}{\name{sorf}}
\newcommand{\cnngprf}{\name{cnn+gp(rf)}}
\newcommand{\cnngpsorf}{\name{cnn+gp(sorf)}}

\title{Calibrating Deep Convolutional Gaussian Processes}

\author{
  Gia-Lac Tran \\
  Department of Data Science\\
  EURECOM, France\\
  \And
  Edwin V.~Bonilla \\
  School of Computer Science and Engineering \\
  University of New South Wales, Australia \\
  \AND
  John P. Cunningham \\
  Department of Statistics \\
  Columbia University, NYC, USA \\
  \And
  Pietro Michiardi \\
  Department of Data Science\\
  EURECOM, France\\
  \And
  Maurizio Filippone \\
  Department of Data Science\\
  EURECOM, France\\
}

\begin{document}

\maketitle

\begin{abstract}
The wide adoption of Convolutional Neural Networks (\cnns) in applications where decision-making under uncertainty is fundamental, has brought a great deal of attention to the ability of these models to accurately quantify the uncertainty in their predictions. Previous work on combining \cnns with Gaussian processes (\gps) has been developed under the assumption that the predictive probabilities of these models are well-calibrated. In this paper we show that, in fact,  current combinations of \cnns and \gps are miscalibrated.  We proposes a novel combination that considerably outperforms previous approaches on this aspect, while achieving state-of-the-art performance on image classification tasks. 


\end{abstract}





\section{Introduction}


The wide adoption of  Convolutional Neural Networks (\cnns)  in increasingly popular pieces of technology such as  self driving cars and medical imaging, where decision-making under uncertainty is fundamental, has brought  attention to the ability of these learning architectures to accurately quantify the uncertainty in their predictions \cite{Kendall17,Gal16b}.
In short, the reliability of predictive probabilities of learning algorithms can be  evaluated  through the analysis of their calibration \cite{Flach16}.  In particular, a classifier is well calibrated when its output offers an accurate account of the probability of a given class, i.e.~when it predicts a given class label with probability $p$ that matches the true proportion $p$ of test points belonging to that class. 

The calibration properties of standard classifiers and  neural networks  have been studied in the literature \cite{Kull17,Niculescu-Mizil05}, which has  shown that classifiers that use the standard cross-entropy loss are generally well calibrated. 
Perhaps surprisingly, modern \cnns, which are a particular case of deep neural networks (\dnns), have been found to be miscalibrated, and the depth of convolutional filters is the main factor affecting calibration \cite{Guo17}. The work in \cite{Guo17} shows that regularization, implemented through weight decay, improves calibration and that, ultimately, simple methods such as post-calibration \cite{Platt99b} can be an effective remedy for most \cnns calibration issues.

Alternatively, Bayesian \cnns \cite{Gal16b} where convolutional filters are inferred using Bayesian inference techniques, seem like perfect candidates to model uncertainty in these architectures in a  principled way.  
However, while Bayesian \cnns have been shown to be effective in obtaining state-of-the-art performance in image classification tasks, we are not aware of studies that show their calibration properties. Hence, our first contribution is to investigate the calibration properties of Bayesian \cnns.

Along a similar vein, independently of the works on Bayesian \cnns, there have been other attempts to give a probabilistic flavor to \cnns by combining them with Gaussian processes (\gps, \cite{Rasmussen06}).
Most of these approaches can be seen as a way to parameterize a \cnn-based covariance for \gps, and the aim is to learn end-to-end both the filters and the \gps (see, e.g., \cite{Bradshaw17,Wilson16}). 
A crucial aspect that the literature has overlooked, however, is that methods that combine \cnns and \gps suffer from the same issues of miscalibration that characterize modern \cnns. 
Therefore, the second contribution of this paper is to show that current combinations of \cnns and \gps are miscalibrated. 

Consequently, as our third contribution, we propose a novel combination of \cnns and \gps that is indeed well-calibrated, while being simple to implement. 
In particular, we propose to replace the fully connected layers of \cnns with \gps that we approximate with random features \cite{Cutajar17,Gredilla10}. 
Due to this approximation, the resulting model becomes a Bayesian \cnn with a nonlinear transformation applied to the convolutional features.
Building on the connection between variational inference and dropout, we apply Monte Carlo dropout (\mcd, \cite{Gal16}) to carry out joint inference over the filters and the approximate \gps, thus obtaining an end-to-end learning method for the proposed model, which we call \cnngprf. 
%
The resulting approach is characterized by a number of attractive features:
(i) it is well calibrated, given that it uses the multinomial likelihood and the filters are regularized using Bayesian inference techniques; 
(ii) it is as scalable as state-of-the-art \cnns, in so much as it can be trained using mini-batch updates and can exploit GPU and distributed computing;
(iii) unlike other works that combine \cnns and \gps, it is as easy to implement as standard \cnns, as it leverages the equivalence of \gps approximated with random features and Bayesian \dnns \cite{Cutajar17,Gal15,Neal96}, and the connections between dropout and variational inference \cite{Gal16}.
We extensively validate these properties in a variety of image classification tasks. 

Our final contribution extends the above framework by replacing the last layer of \cnns with Deep \gps \cite{Cutajar17} and use structured random features to obtain faster and more compact \gp approximations \cite{Le13,Yu16}.
In all, our proposal considerably improves on classification accuracy compared to previous combinations of \cnns and \gps (e.g., $\sim$$89\%$ on \cifart and $\sim$$75\%$ on \cifarh, all without data augmentation), while being competitive with state-of-the-art \cnns; we are not aware of other \gp works that approach these results.
Crucially, we achieve these performance without compromising on calibration, again considerably improving on previous approaches that combine \cnns and \gps.

\section{Related Work}

\paragraph{Calibration of Convolutional Networks:}
The issue of calibration of classifiers in machine learning was popularized in the 90's with the use of support vector machines for probabilistic classification \cite{Platt99b}.
Calibration techniques aim to learn a transformation of the output using a validation set in order for the transformed output to give a reliable account of the actual probability of class labels \cite{Flach16}; interestingly, calibration can be applied regardless of the probabilistic nature of the untransformed output of the classifier.
Popular calibration techniques include Platt scaling \cite{Platt99b} and isotonic regression \cite{Zadrozny02}.

Classifiers based on Deep Neural Networks (\dnns) have been shown to be well-calibrated \cite{Niculescu-Mizil05}. 
The reason is that the optimization of the cross-entropy loss promotes calibrated output. 
The same loss is used in Platt scaling and it corresponds to the correct multinomial likelihood for class labels.
Recent sudies on the calibration of \cnns, which are a particular case of \dnns, however, show that depth has a negative impact on calibration, despite the use of a cross-entropy loss, and that regularization improves the calibration properties of classifiers \cite{Guo17}.

\paragraph{Combinations of Conv Nets and Gaussian Processes:}
Thinking of Bayesian priors as a form of regularization, it is natural to assume that Bayesian \cnns  can ``cure'' the miscalibration of modern \cnns.
Despite the abundant literature on Bayesian \dnns \cite{Neal96,Mackay94}, far less attention has been devoted to Bayesian \cnns~\cite{Gal16}, and the calibration properties of these approaches have not been investigated. 

Several approaches have  proposed the combination of \cnns and \gps as a means to give a probabilistic character to \cnns. 
Most of these works are based on ideas developed in the context of manifold \gps \cite{Calandra16}, where inputs are transformed using some parametric transformation of the input.
In these works, the parametric transformation is based on convolutional layers, and scalability to large data 
is achieved through the use of ideas drawn from the literature on scalable \gps, for example the Stochastic Variational Deep Kernel Learning (\svdkl) approach in  \cite{Wilson16}.
In contrast, the work on hybrid \gps and \dnns (\gpdnn, \cite{Bradshaw17}) combines \cnns and \gps using an inducing point approximation. 
Other recent approaches that aim to introduce convolutions in the calculation of the covariance between images include \cite{VanDerWilk17}, which proposes a way to construct covariances between domains/patches, mimicking the computations in \cnns. 

In this work, we propose an alternative way to combine \cnns and \gps, where \gps are approximated using random features expansions \cite{Rahimi08,Gredilla10}.
The random feature expansion approximation amounts in replacing the orginal kernel matrix with a low-rank approximation, turning \gps into Bayesian linear models.
Combining this with \cnns leads to a particular form of Bayesian \cnns, much like \gps and \dgps are particular forms of Bayesian \dnns \cite{Duvenaud14,Gal16,Neal96}.
Inference in Bayesian \cnns is intractable and requires some form of approximation. 
In this work, we draw on the interpretation of dropout as variational inference, employing the so-called Monte Carlo Dropout (\mcd, \cite{Gal16}) to obtain a practical way of combining \cnns and \gps.

\section{Combinations of \cnns and \gps are miscalibrated}

Consider a $\nclass$-class image classification task where $\Xmat$ denotes a set of $\nobs$ images $\xvect_i \in \mathbb{R}^{p_x \times p_y} (1\leq i \leq n)$, and  $\Ymat$ is the matrix consisting of the corresponding one-hot encoded labels $\yvect_i$ stacked by row.
We can use various metrics to determine the quality of a classifier, and here we focus in particular on calibration.
Let $\gvect(\xvect)$ be the output of a classifier for an input image $\xvect$.
To compute the calibration properties of a classifier, consider a partitioning of the test set $\Xmat_*$ into disjoint sets $\{\Xmat_{1}, \dots, \Xmat_{M}\}$, such that each subset $\Xmat_{m}$ contains the inputs yielding predictions in the range $(\frac{m-1}{M}, \frac{m}{M}]$. 
Hence, the confidence associated with each subset $\Xmat_m$ is characterized by the midpoint of its corresponding range, i.e.\ $\text{confidence}(\Xmat_m) = \frac{m-0.5}{M}$. 
Then, the accuracy for each subset can be evaluated as follows: 
\begin{equation}
\text{accuracy}(\Xmat_m) = \frac{1}{|\Xmat_m|} \sum_{\xvect_* \in \Xmat_m} \delta\left(\arg\max(\yvect_*) - \arg\max(\gvect(\xvect_*)) \right) \text{,}
\label{eq:accuracy}
\end{equation}
where $\delta(x)$ is equal to one if $x = 0$, and zero otherwise. 

In what follows, we use reliability diagrams to assess calibration, where we plot accuracy as a function of confidence for the subsets $\{\Xmat_{1}, \dots, \Xmat_{M}\}$.
For a perfectly calibrated classifier, we expect $\text{accuracy}(\Xmat_m) = \text{confidence}(\Xmat_m)$ for all $m$, with deviations implying that the class probabilities are either underestimated or overestimated.
A useful summary statistics that can be extracted from reliability diagrams is the 
 \emph{Expected Calibration Error} (\ece), which is the average of the absolute difference between accuracy and confidence \emph{weighted} according to its size:
\begin{equation}
\ece = \sum_{m=1}^{M} \frac{|\Xmat_m|}{|\Xmat_*|} \left\vert\text{accuracy}(\Xmat_m) - \text{confidence}(\Xmat_m)\right\vert \text{.}
\end{equation}
Another metric that measures the accuracy in predicting class probabilities is the \brier score, defined as the squared distance between labels and outputs averaged across classes and test points:
\begin{equation}
\brier = \frac{1}{N_{\mathrm{test}}} \sum_{\xvect_* \in \Xmat_*} \frac{1}{\nclass} \sum_{k=1}^{\nclass} \left((\yvect_*)_k - (\gvect(\xvect_*))_k \right)^2
\end{equation}

\def \offset {6}

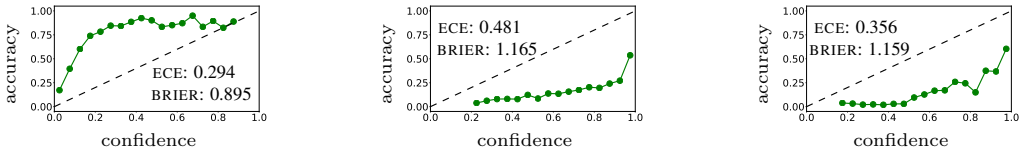
\begin{figure}[t]
\vskip 0.2in
\begin{center}
\begin{tikzpicture}

\node[inner sep=0pt](mnist-lenet-mnll) at (2.5, 5){\includegraphics[width=3.2cm]{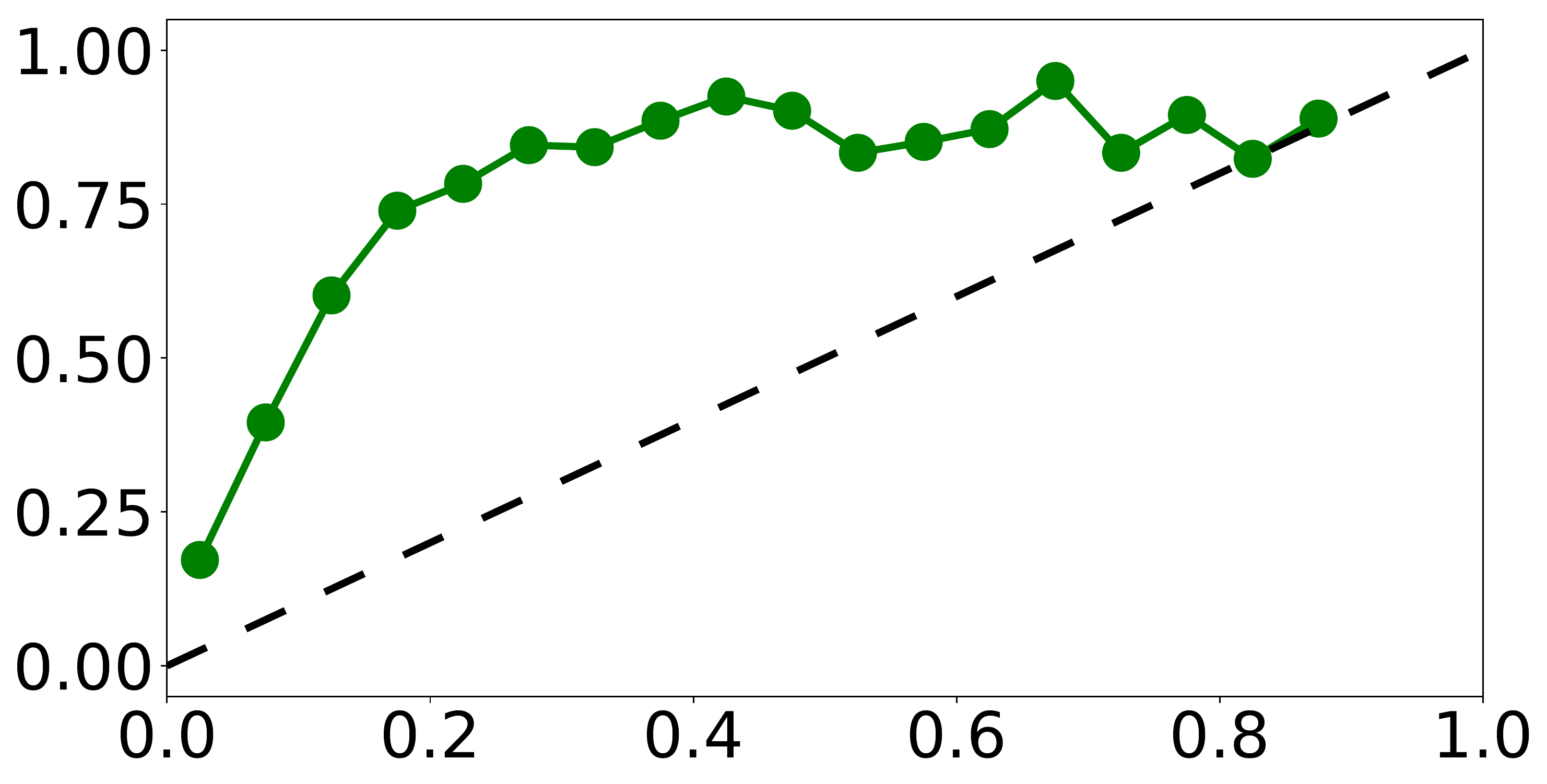}};

\node[inner sep=0pt](mnist-lenet-ece) at (7.5, 5){\includegraphics[width=3.2cm]{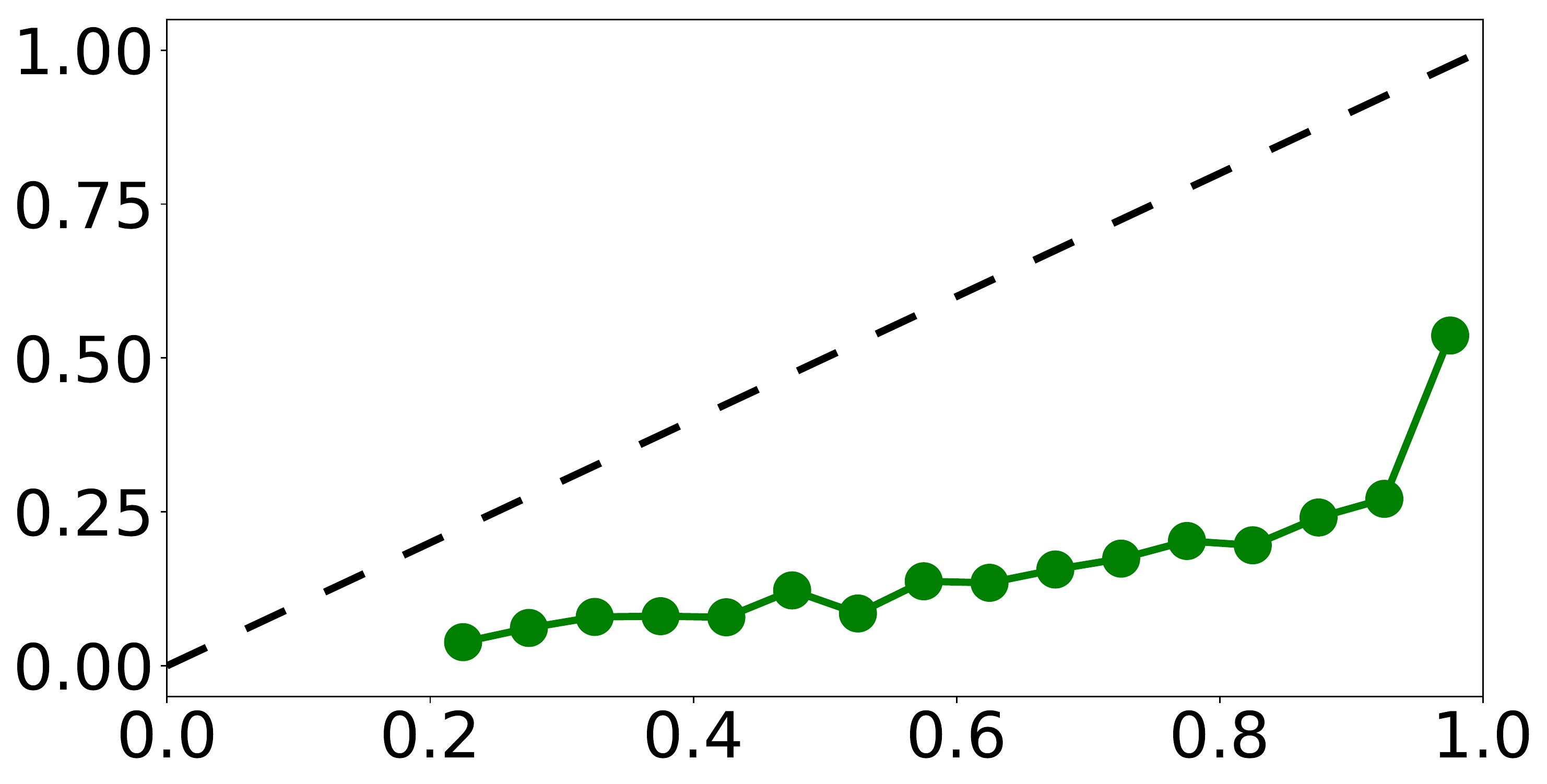}};

\node[inner sep=0pt](mnist-lenet-ece) at (12.5, 5){\includegraphics[width=3.2cm]{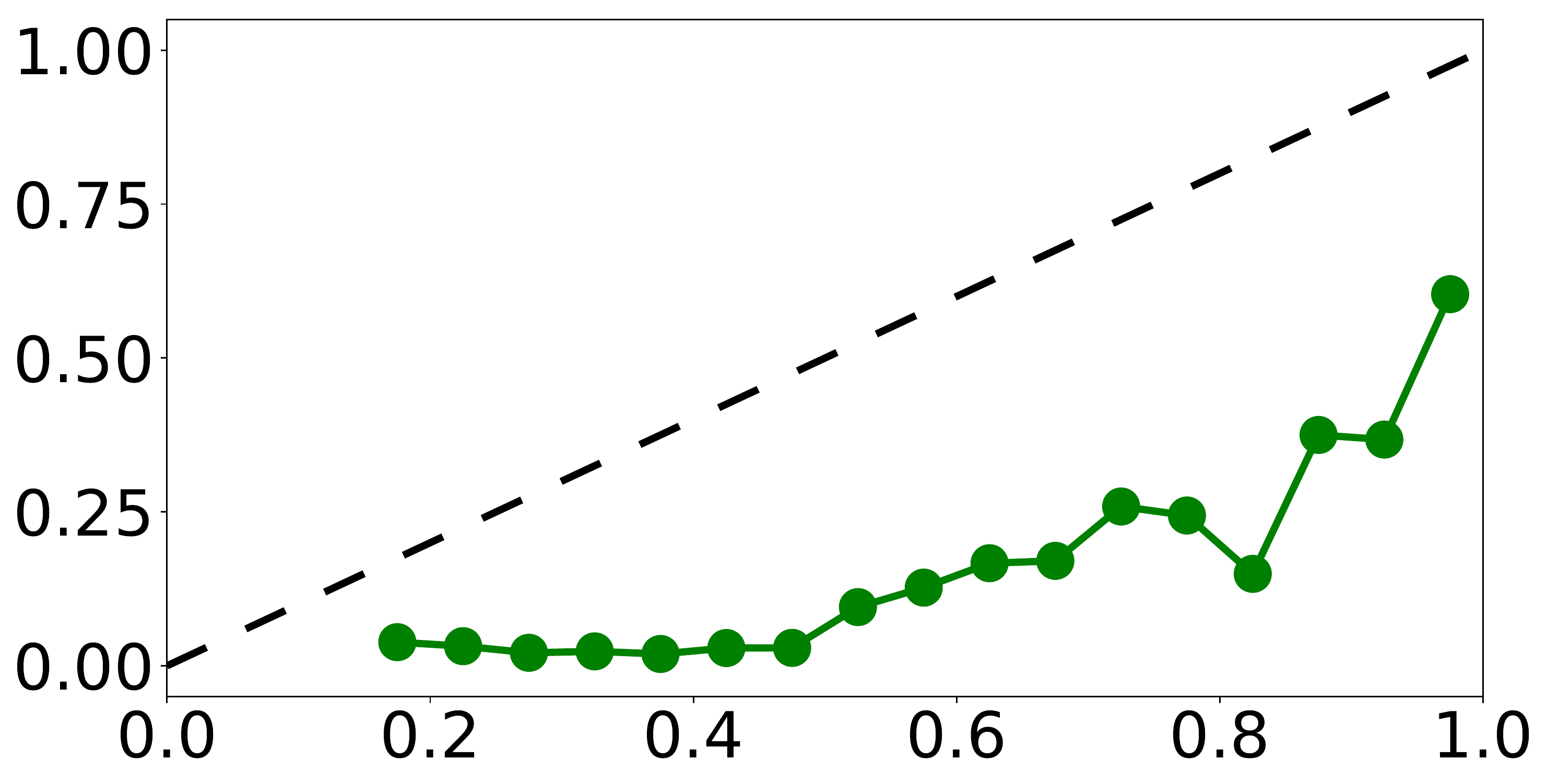}};

\node(text) at (2.5, 4.0) {\scriptsize $\mathrm{confidence}$};
\node(text) at (7.5, 4.0) {\scriptsize $\mathrm{confidence}$};
\node(text) at (12.5, 4.0) {\scriptsize $\mathrm{confidence}$};

\node[rotate=90](text) at (0.7, 5) {\scriptsize $\mathrm{accuracy}$};
\node[rotate=90](text) at (5.7, 5) {\scriptsize $\mathrm{accuracy}$};
\node[rotate=90](text) at (10.7,5) {\scriptsize $\mathrm{accuracy}$};

\node(text) at (3.1, 4.9) { \scriptsize \ece: 0.294};
\node(text) at (3.2, 4.6) { \scriptsize \brier: 0.895};

\node(text) at (6.9, 5.5) { \scriptsize \ece: 0.481};
\node(text) at (7.0, 5.2) { \scriptsize \brier: 1.165};

\node(text) at (11.9, 5.5) { \scriptsize \ece: 0.356};
\node(text) at (11.95, 5.2) { \scriptsize \brier: 1.159};

\end{tikzpicture}
\caption{Reliability diagrams for three state-of-the-art combinations of \cnns and \gps on \cifarh with the \resnet convolutional structure. 
Left: \gpdnn \cite{Bradshaw17}.
Center: \cgp \cite{VanDerWilk17}. 
Right: \svdkl \cite{Wilson16}. 
See table~\ref{tab:configurations:cnns} for details on the \resnet convolutional architecture that we apply to \cifarh.
}
\label{reliability diagrams}
\end{center}
\vskip -0.2in
\end{figure}


In figure~\ref{reliability diagrams}, we report the reliability diagrams of three state-of-the-art combinations of \cnns and \gps applied to the \cifarh data set, along with the corresponding values of \ece and \brier scores. 
The left and right panels of the figure show the \gpdnn approach in \cite{Bradshaw17} and \svdkl in \cite{Wilson16}, respectively, where we use a \resnet architecture for the convolutional layers.
The central panel of the figure reports the reliability diagram for the \cgp in \cite{VanDerWilk17} where there is no equivalent of a \cnn architecture.

The results indicate that current approaches that combine \cnns and \gps are miscalibrated, with a tendence of being either underconfident (\gpdnn) or overconfident (\cgp and \svdkl) in predictions.
This is an important and perhaps surprising finding, because one of the motivations to combine \cnns with \gps is to do better quantification of uncertainty compared to plain \cnns. 
In the experiments section we report more extensively on the calibration of these classifiers, as well as illustrating other performance metrics.
These considerations call for the study of better ways to combine \cnns and \gps to recover calibration while attempting to improve on standard 
metrics such as error rate and test log-likelihood. 
The next section illustrates our proposal that achieves this goal.


\section{\cnngprf: Conv Nets with Random Feature Expanded \gps}

In the proposed model, the labels $\Ymat_{i \cdot}$ are assumed to be conditionally independent given a set of corresponding latent variables $\Fmat_{i \cdot}$, i.e.~we consider the likelihood 
$
p(\Ymat | \Fmat) = \prod_{i=1}^\nobs p(\Ymat_{i \cdot}\ | \Fmat_{i \cdot}) \text{,}
$
where the latent variables $\Fmat$ are  realizations of a set of $\nclass$ functions $f_j(\mathbf{x})$ at the input images $\xvect_1, \dots, \mathbf{x}_n$, i.e., $(\Fmat)_{ij} = f_j(\xvect_i)$ for $j=1, \ldots, \nclass$. 
In this work we focus on functions $f_j(\mathbf{x})$ that are modeled using \gps; note that extension to \dgps is actually easy to consider in our framework, and we will report results on this choice in the experiments. 
Each individual $p(\Ymat_{i \cdot}\ | \Fmat_{i \cdot})$ is multinomial with probabilities obtained using a softmax transformation of the latent variables. 

Due to the \gp modeling assumption, the latent function values $\Fmat_{\cdot j} = (f_j(\xvect_1), \ldots, f_j(\xvect_n))^{\top}$ are jointly Gaussian with $p(\Fmat_{\cdot j} | \Xmat, \boldsymbol{\theta})  \sim {\cal N} (\mathbf{0},\mathbf{K})$, where $\mathbf{K}$ is the covariance matrix.
The entries of the covariance matrix $\mathbf{K} = \left\{k\left(\mathbf{x}_i, \mathbf{x}_j | \thetavect\right)\right\}_{i,j}$, are specified by a covariance (kernel) function $k$ (with hyperparameters $\thetavect$) and, this form is shared across output dimensions, although this can be relaxed and allow for a different $k$ for the $Q$ outputs.

Instead of applying the \gp modeling directly to the images, we propose to employ a transformation $\cvect(\xvect | \convweights)$ using convolutional layers,
where $\convweights$ denotes the parameters of these convolutional layers.  
The vector-valued function $\cvect(\xvect | \convweights)$ is differentiable as it implements a series of differentiable operations, such as convolutions and pooling.
This is one of the key successes of \cnn models that allows for the learning of their filters, which we exploit for the end-to-end learning of our model.

Inference in this model requires being able to characterize the posterior over all or a selected group of model parameters $\thetavect, \convweights$, but this posterior is analytically intractable and thus computationally prohibitive \cite{Rasmussen06}.
In the remainder of this paper, we will build on previous work on scalable inference for \gps and \dgps with random features \cite{Cutajar17} to obtain an approximation to the proposed model that can be learned end-to-end.
\subsection{Random Feature Expansions for Gaussian Processes}
Na\"{i}ve inference in \gp models requires algebraic operations with $\mathbf{K}$ that would cost $\bigO(n^3)$ in time. 
Popular approaches to recover tractability use low-rank approximations of the kernel matrix.  
Among this family of low-rank approximations, we choose to work with random feature approximations \cite{Gredilla10,Cutajar17}.
The reason is that they offer a number of possible extensions to speedup computations (e.g., using structured approximations \cite{Le13,Yu16}) and increase the complexity of the model (e.g., considering Deep \gps \cite{Cutajar17}); we will elaborate on this in the experiments section.
In random feature expansions, the kernel matrix is replaced by a low-rank approximation
$\mathbf{K} \approx \Phimat \Phimat^{\top}$,
with $\Phimat \in \mathbb{R}^{n \times m}$ and $m \ll n$. 
This approximation suggests the construction of a Bayesian linear model to approximate the \gp latent variables as $\Fmat = \Phimat \Wmat$.
Using $p(W_{ij}) = \norm(W_{ij} | 0, 1)$ it is straightforward to show that the covariance of each of the latent functions $\Fmat_{\cdot j}$ is indeed the approximate $\mathbf{K}$, as $\mathrm{cov}(\Fmat_{\cdot j}) = \E(\Phimat \Wmat_{\cdot j} \Wmat_{\cdot j}^{\top} \Phimat^{\top}) = \Phimat \E(\Wmat_{\cdot j} \Wmat_{\cdot j}^{\top}) \Phimat^{\top} = \Phimat \Phimat^{\top} \approx \mathbf{K}$.

In this work, we focus in particular on the order-one \arccosine kernel \cite{Cho09}
\begin{equation} \label{eq:covariance:arccosine:ard}
k_{\text{arc}}^{(1)}(\mathbf{x}_i, \mathbf{x}_j | \convweights, \thetavect) = 
\frac{\sigma^2}{\pi} \left\| \Lambda^{-\frac{1}{2}} \cvect(\xvect_i | \convweights) \right\| \left\| \Lambda^{-\frac{1}{2}} \cvect(\xvect_j | \convweights) \right\| \left[ \sin(\alpha) + (\pi - \alpha) \cos(\alpha)  \right] \text{,}
\end{equation}
where $\thetavect =  (\sigma, \Lambdamat=\diag(\ell_1^2, \ldots, \ell_\nconv^2))$ and $\alpha$ is the angle between $\Lambda^{-\frac{1}{2}} \cvect(\xvect_i | \convweights)$ and $\Lambda^{-\frac{1}{2}} \cvect(\xvect_j | \convweights)$.

%
The \arccosine covariance has a convenient integral representation
that allows for a Monte Carlo approximation, obtaining a low-rank approximation to the covariance matrix involving Rectified Linear Unit (\relu) activations \cite{Cho09}
\begin{equation} \label{eq:random:feat:expansion:arccosine}
        \Phimat_{\text{arc}} = \sqrt{\frac{ 2 \sigma^2}{N_{\mathrm{RF}}}}  \max\left(\zerovect, \Cmat(\Xmat | \convweights) \, \Omegamat\right)  \text{.}
\end{equation}
In this expression, we have defined $\Cmat(\Xmat | \convweights)$ as the matrix resulting from the application of convolutional layers to the image training set $\Xmat$ and $\Omegamat$ is obtained by stacking $N_{\mathrm{RF}}$ samples from $p(\omegavect) = \norm\left(\omegavect | \zerovect, \Lambdamat^{-1} \right)$ by column. 
Note that in the case of a popular Radial Basis Function (\rbf) covariance, it is possible to obtain a similar random feature approximation, where the \relu activation is replaced by trigonometric functions; see \cite{Rahimi08} and the supplement for details. 

\subsection{End-to-end learning of the proposed \cnngprf model}



Inference in the proposed model is intractable due to the likelihood that is not conjugate to the \gp prior. 
Further complications stem from the need to infer kernel parameters, which include convolutional parameters, and the need to be able to scale to large data. 
Our aim is to carry out inference within a consistent framework that is characterized by simplicity, as described in what follows. 

We start by introducing an approximate posterior over $\Wmat$ and $\convweights$, that we denote as $q(\Wmat, \convweights)$.
Following standard variational inference arguments, we can define an operative way to obtain these approximate posteriors. 
The log-marginal likelihood $\LL = \log \left[p(\Ymat | \Xmat, \Omegamat, \thetavect\right]$ can be bounded by the sum of an expected log-likelihood term and a negative Kullback-Leibler (KL) divergence term: 
\begin{equation}
\label{eq:elbo}
\LL \geq 
\E_{q(\Wmat, \convweights)} \left( \log\left[ p\left(\Ymat | \Xmat, \Wmat, \convweights, \Omegamat, \thetavect\right) \right] \right)
- \mathrm{KL}\left[q\left(\Wmat, \convweights\right) \| p\left(\Wmat, \convweights\right) \right].
\end{equation}
Variational inference amounts to optimizing the lower bound above with respect to $q(\Wmat, \convweights)$ and any other parameters of interest. 
Here we assume $\Omegamat$ fixed from the prior, while we wish to optimize its prior parameters $\thetavect$, but note that $\Omegamat$ could also be inferred variationally (see supplement and \cite{Gal15,Cutajar17}).

We have now a number of options on the form for the approximate posteriors $q(\Wmat, \convweights)$. 
In previous works on variational inference for \dnns, it has been proposed to define the approximating distributions to be Gaussian and factorized across parameters \cite{Kingma14,Graves11}. 
The drawback of this is that it doubles the number of parameters. 
Alternatively, we can rely on the connections between dropout and variational inference \cite{Gal16,Gal16b} to obtain an easier approximate inference scheme, which is also known as Monte Carlo Dropout (\mcd).
Focusing on the weights for now, the connection with dropout is apparent if we rewrite
\begin{equation}
\Wmat = \Mmat_w \, \diag[\zvect_w] \qquad \mathrm{with} \qquad (\zvect_w)_i \sim \mathrm{Bernoulli}(\pi_w) \text{.}
\end{equation}
The reparameterization introduces variational parameters $\Mmat_w$ (one for each weight in $\Wmat$) and a vector of binary variables that can switch on or off the columns of the weight matrix with probability $\pi_w$.
A similar reprameterization can be done for the convolutional parameters $\convweights$, introducing $\Mmat_{\psi}$ and $\zvect_{\psi}$.
The optimization of the lower bound wrt all variational parameters requires being able to evaluate the expectation and the KL term.

The KL term can be approximated following \cite{Gal16}, obtaining a regularization term involving the squared-norm of the parameters
\begin{equation}
\mathrm{KL}\left[q\left(\Wmat, \convweights\right) \| p\left(\Wmat, \convweights\right) \right]
\approx
\frac{\pi_w}{2} \left\| \Mmat_w \right\|^2 + 
\frac{\pi_\psi}{2} \left\| \Mmat_\psi \right\|^2
\end{equation}
The expectation, instead, can be unbiasedly estimated using Monte Carlo and also considering a mini-batch of size $m$:
\begin{equation}
\E_{q(\Wmat, \convweights)} \left( \log\left[ p\left(\Ymat | \Xmat, \Wmat, \convweights, \Omegamat, \thetavect\right) \right] \right)
\approx
\frac{\nobs}{m} \frac{1}{N_{\mathrm{MC}}} \sum_{i = 1}^{N_{\mathrm{MC}}} \sum_{k \in \mathcal{I}_{m}} \log\left[ p\left(\yvect_k | \xvect_k, \Wmat^{(i)}, \convweights^{(i)}, \Omegamat, \thetavect\right) \right] \text{,}
\end{equation}
with $\Wmat^{(i)}, \convweights^{(i)} \sim q(\Wmat, \convweights)$, and $\mathcal{I}_m$ is a set of $m$ indices to select a mini-batch of training points \cite{Kingma14,Graves11}.
This doubly-stochastic approximation is differentiable wrt variational parameters when the Bernoulli variables are fixed.

The approximate objective can now be optimized in the same vein as in standard back-propagation with dropout, noting that dropout is applied to $\Wmat$ as well as convolutional parameters $\convweights$.
What changes, however, is the interpretation of the procedure as stochastic variational inference, whereby the Bernoulli variables are resampled at each iteration. 
A practical implication is in the way we compute the predictive distribution, which has a probabilistic flavor as follows: 
\begin{equation}
p(\yvect_* | \xvect_*, X, \Omegamat, \thetavect) \approx \int p(\yvect_* | \Wmat, \convweights, \xvect_*, X, \Omegamat, \thetavect) q(\Wmat,\convweights) d\Wmat d\convweights \text{,}
\end{equation}
and can be approximated using Monte Carlo by resampling the Bernoulli variables. 
While \mcd has been proposed for \cnns in \cite{Gal16b}, in this work we extend it to the case of joint inference over convolutional parameters and the \gp approximation in the \cnngprf model, thus obtaining a  practical inference and prediction scheme combining \cnns and \gps.

\subsection{Extensions}

\paragraph{Structured random feature approximations:}

One of the advantages of the proposed model, compared to other \gp approximations, is that it can exploit structured random feature expansions to accelerate computations and reduce the size of the approximate \gp \cite{Le13,Yu16}. 
In the random features approximation, random features are constructed by multiplying $\Omegamat$ with the convolutional features.
Without loss of generality, assuming that $\Omegamat \in \mathbb{R}^{ \nbasis \times \nconv}$ and $\cvect(\xvect | \convweights) \in\mathbb{R}^{\nconv \times 1}$,
the cost of computing products $\Omegamat \cvect(\xvect | \convweights)$ is $\mathcal{O}\left( \nbasis \nconv \right)$, while storing $\Omegamat$ requires $\mathcal{O}\left(\nbasis \nconv\right)$ storage. 

Structured approximations aim to reduce the time complexity to $\mathcal{O}\left( \nbasis \log \nconv \right)$ and the storage cost to $\mathcal{O}\left(\nbasis + \nconv\right)$.  
Taking a standard random features expansion of the isotropic covariance in equation~\eqref{eq:random:feat:expansion:arccosine} with $\Lambdamat = \ell^{-2} \Imat$ as an example, 
$\Omegamat = \frac{1}{\ell}\Gmat$, 
with $\Gmat_{ij} \sim \mathcal{N}\left(0, 1\right)$.
One way to make computations cheaper is to replace the Gaussian matrix $\Gmat$ with a pseudo-random alternative.
The Structured Orthogonal Random Feature (\sorf) approximation \cite{Yu16} approximates $\Gmat$ through a series of Hadamard transformations of diagonal matrices $\Dmat_i$ with elements randomly sampled from $\{-1, +1\}$, that is $\Gmat \approx \sqrt{d} \Hmat \Dmat_1 \Hmat \Dmat_2 \Hmat \Dmat_3$.
We refer to this variation of the model as \cnngpsorf.
\paragraph{Convolutional Networks with Random-Feature-Expanded Deep \gps:}
A \dgp model represents a deep probabilistic nonparametric approach where the output of one \gp at each layer is used as the input to the \gp in the next layer \cite{Damianou13}. 
Extending the random feature approximation to \dgps and the inference scheme presented here is straightforward; see \cite{Cutajar17} for details. 
The random feature approximation turns the \dgp into a Bayesian \dnn for which we can apply stochastic variational inference to infer model parameters. 
In the experiments section, we explore the possibility to stack a \dgp on top of convolutional layers, and we show the impact of depth on performance. 

\section{Experiments}

\begin{table}[t]
  \caption{\cnn architectures considered in this work. The same architectures are used in \gpdnn and \svdkl by replacing the fully connected layers with \gps, while \cgp does not explicitly use a convolutional structure.}  
  \label{tab:configurations:cnns}
  \centering
  \begin{tabular}{lllll}
    \toprule
    Depth  & Data set     & \cnn architecture & \cnn name & \# Conv features \\
    \midrule
    Shallow & \mnist &  2 Conv Layers + 2 Fully connected & \lenet  & 4096   \\
    Shallow & \cifart  & 2 Conv Layers + 3 Fully connected & \lenet  & 4096  \\
    Deep & \cifart     & 30 Conv Layers + 1 Fully connected & \resnet & 64   \\
    Deep & \cifarh     & 150 Conv Layers + 1 Fully connected & \resnet &  64   \\
    \bottomrule
  \end{tabular}
\end{table}




We carry out the experimental evaluation using popular benchmark datasets, such as \mnist, \cifart and \cifarh and with a number of popular \cnn architectures based on \name{LeNet} and \name{resnet} (see table~\ref{tab:configurations:cnns}). 
We report three state-of-the-art competitors combining \cnns and \gps, namely \gpdnn \cite{Bradshaw17}, \svdkl \cite{Wilson16}, and \cgp \cite{VanDerWilk17}.
We also report Bayesian \cnns, as suggested in \cite{Gal16b} and \cnns with post-calibration as proposed in \cite{Guo17}, which we refer to as \name{cnn+mcd} and \name{cnn+cal}, respectively. 
For all the competing methods we used available implementations, adding the same \cnn architecture to ensure a fair comparison.
In all experiments, we use a batch-size $m = 1000$ and the Adam optimizer with default learning rate \cite{Kingma14b}. 
In the methods that use \mcd, we use a dropout rate of $0.5$ for all parameters.

The results are reported in figure~\ref{model comparison}, where we show results for different training sizes $\nobs$, keeping the classes balanced.
In the figure, we report the calibration measures that we have introduced earlier, namely \ece and \brier scores, and we also report the classification error rate (\err) and the mean negative test log-likelihood (\mnll).
Compared to other combinations of \cnns and \gps, our proposal improves considerably on all metrics. 
It is interesting to see that our proposal is competitive with Bayesian \cnns employing \mcd, with only a marginal improvement on \err and \mnll in some configurations. 
Compared to plain \cnns with post-calibration, our proposal is considerably better, although in some configurations the former is superior in \ece and \brier; this is expected given that this is the metric that is optimized on a validation set. 

The two variants of our approach, namely \cnngprf where we learn the frequencies $\Omegamat$ and \cnngpsorf where we sample $\Omegamat$ from its prior, are comparable.
This suggests that the extra level of complexity of learning the spectral frequencies does not lead to substantial gains in performance and that the structured random feature approximation yields satisfactory performance.
We also note that these results have been obtained by fixing the covariance parameters $\thetavect$ of the \gp, as we found it to be unstable when learning these jointly with $\Omegamat$. 
This might be the reason why these parameters were learned through cross-validation in \cite{Gal17}. 
In the supplement, we report the results obtained when learning $\thetavect$, which we found yielding similar performance as fixing them. 
All these observations corroborate the hypothesis that most of the performance of \cnn-based classification models is due to the convolutional layers.

In summary, figure~\ref{model comparison} shows that our \cnngprf is the best strategy for calibrating these models compared to other approaches using \gps. 
Furthermore, we found perhaps surprisingly that \mcd has similarly performance.  
In the supplementary material, we report results on \gpdnn where we infer convolutional parameters using \mcd, so as to gain insights as to whether most of the improvements in performance are due to this form of regularization.
The results support the intuition that inferring these parameters yields improvements in calibration, but also that our \cnngprf still offers better performance.

\def \offset {4.5}

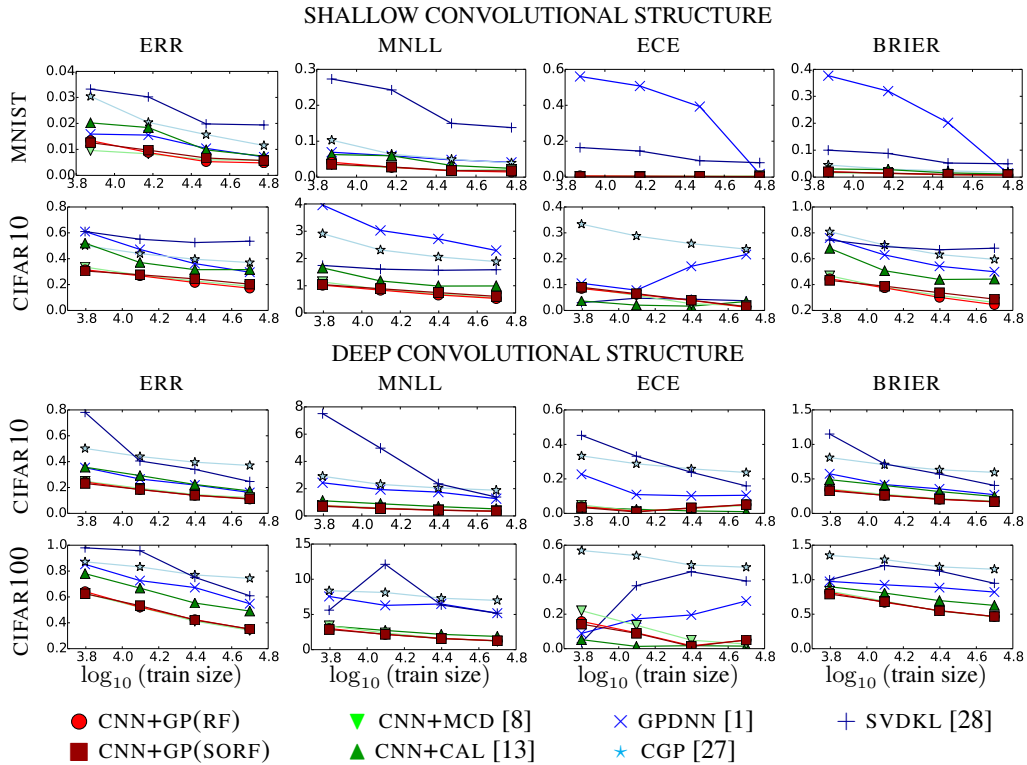
\begin{figure}[t]
\vskip 0.2in
\begin{center}
\begin{tikzpicture}
\node(mnist-text) at (7, 14.8 + \offset) {\small SHALLOW CONVOLUTIONAL STRUCTURE};

\node(mnist-err-text) at (2.0, 14.4 + \offset) {\err};
\node(mnist-mnll-text) at (5.3, 14.4 + \offset) {\mnll};
\node(cifar10-err-text) at (8.6, 14.4 + \offset) {\ece};
\node(cifar10-mnll-text) at (11.9, 14.4 + \offset) {\brier};

\node[rotate=90](text) at (0.1, 13.3 + \offset) {\mnist};
\node[rotate=90](text) at (0.1, 11.5 + \offset) {\cifart};

\node[inner sep=0pt](mnist-lenet-err) at (2.0, 13.3 + \offset){\includegraphics[width=3.2cm]{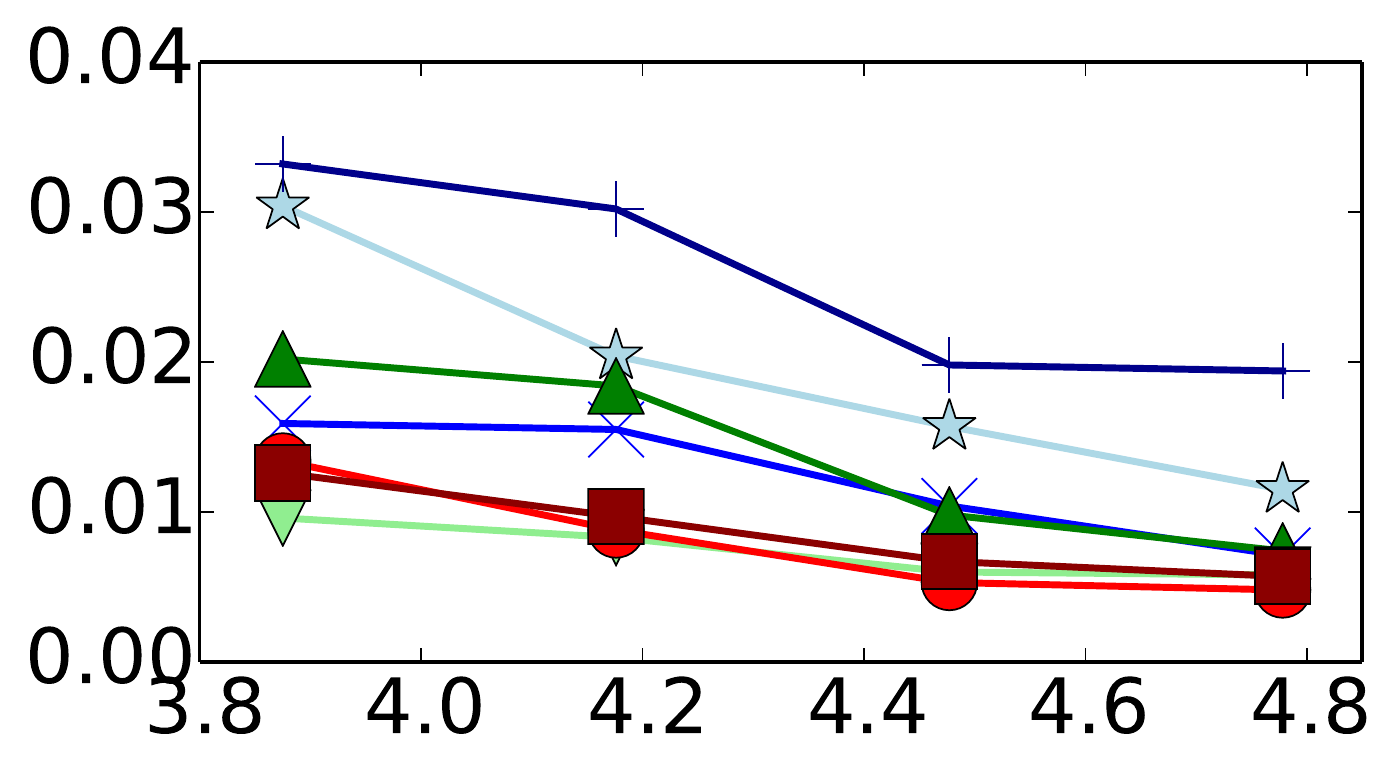}};
\node[inner sep=0pt](mnist-lenet-mnll) at (5.3, 13.3 + \offset){\includegraphics[width=3.2cm]{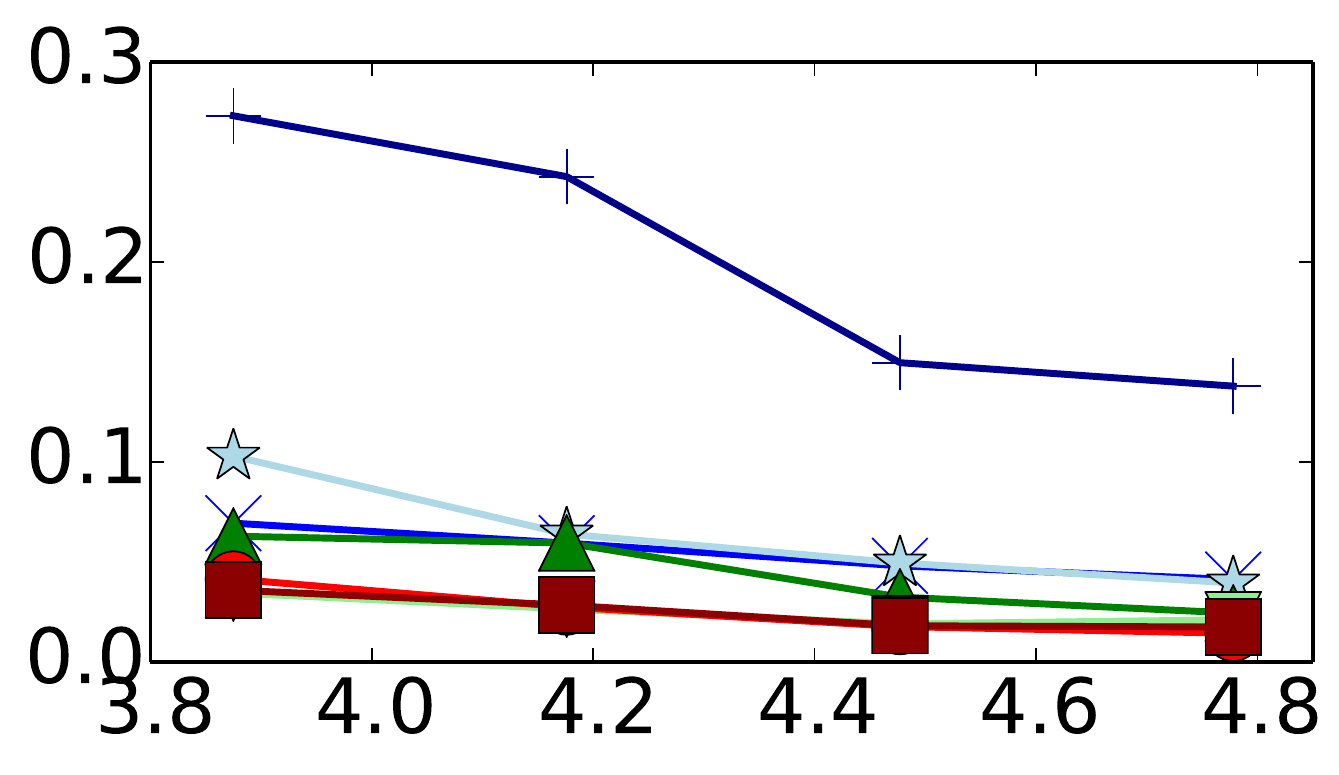}};
\node[inner sep=0pt](mnist-lenet-ece) at (8.6, 13.3 + \offset){\includegraphics[width=3.2cm]{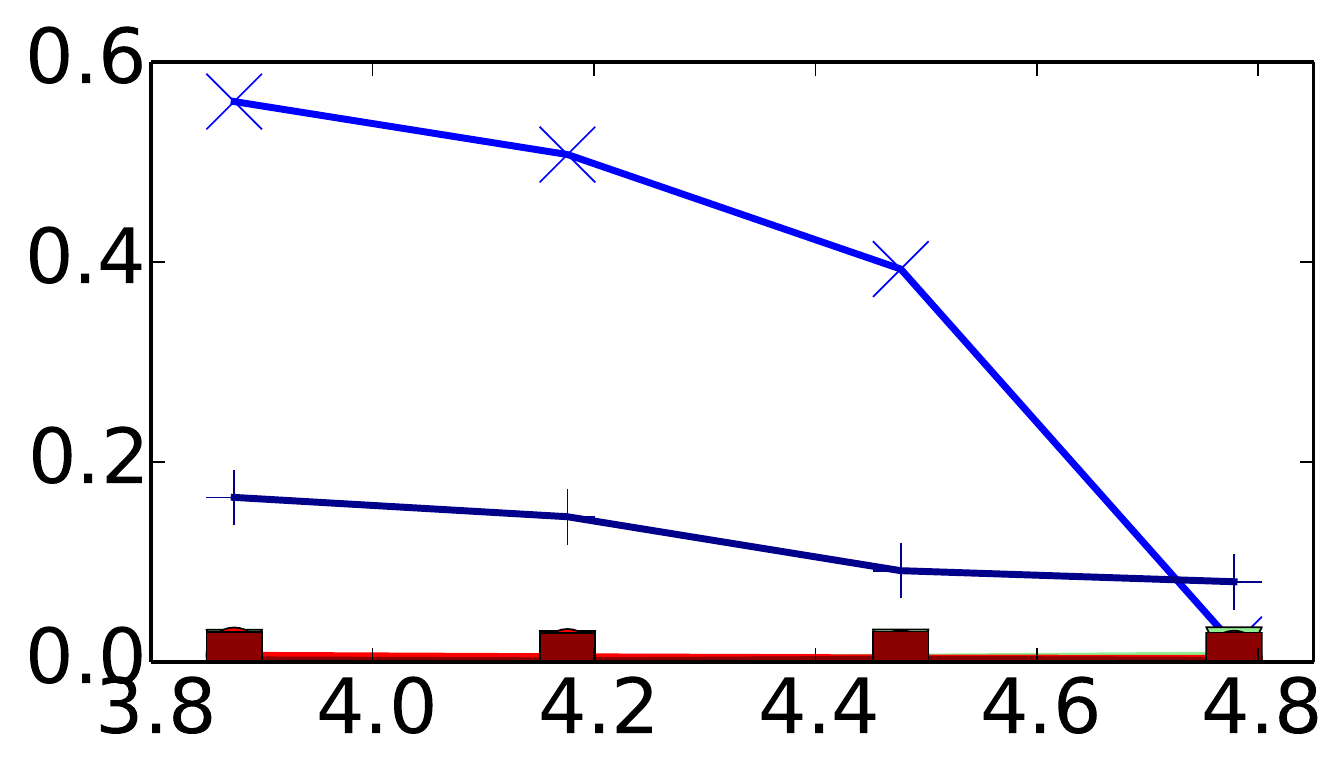}};
\node[inner sep=0pt](mnist-lenet-brier) at (11.9, 13.3 + \offset){\includegraphics[width=3.2cm]{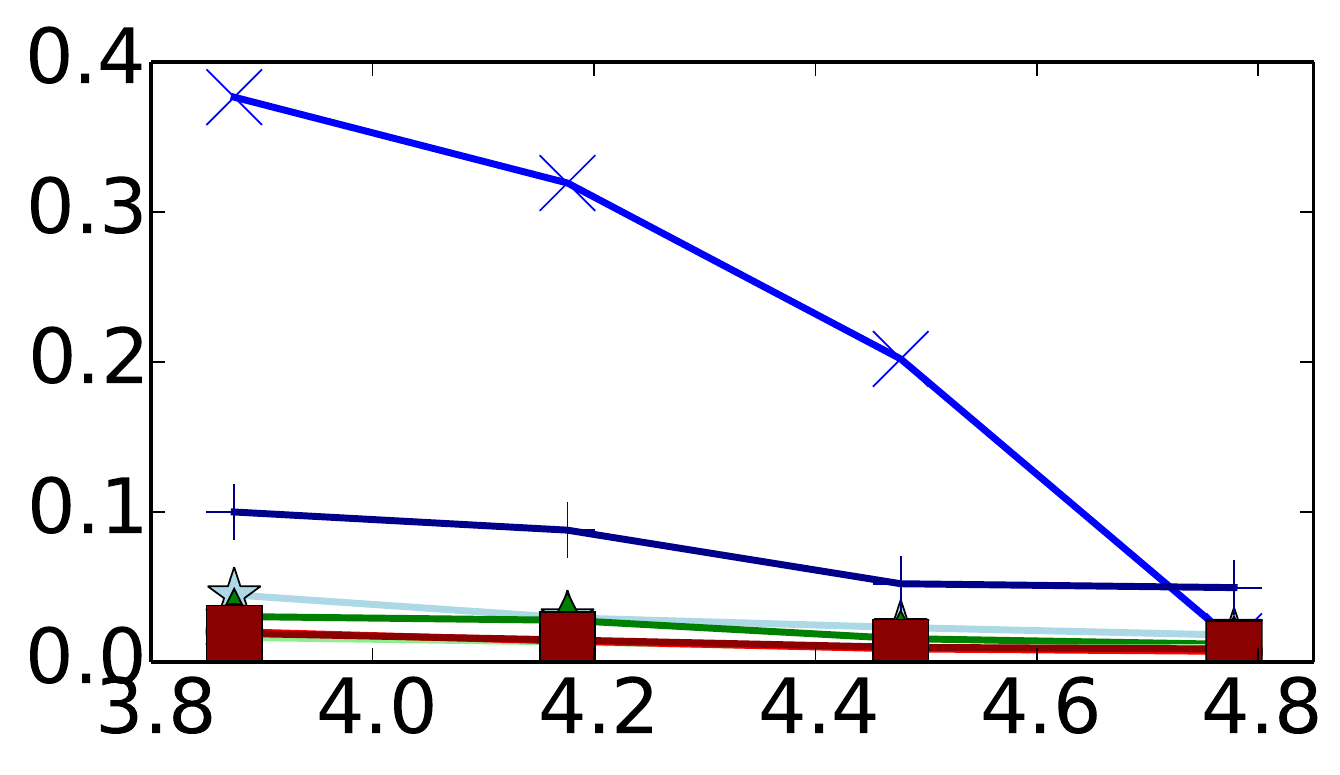}};

\node[inner sep=0pt](cifar10-lenet-err) at (2.0, 11.5 + \offset){\includegraphics[width=3.2cm]{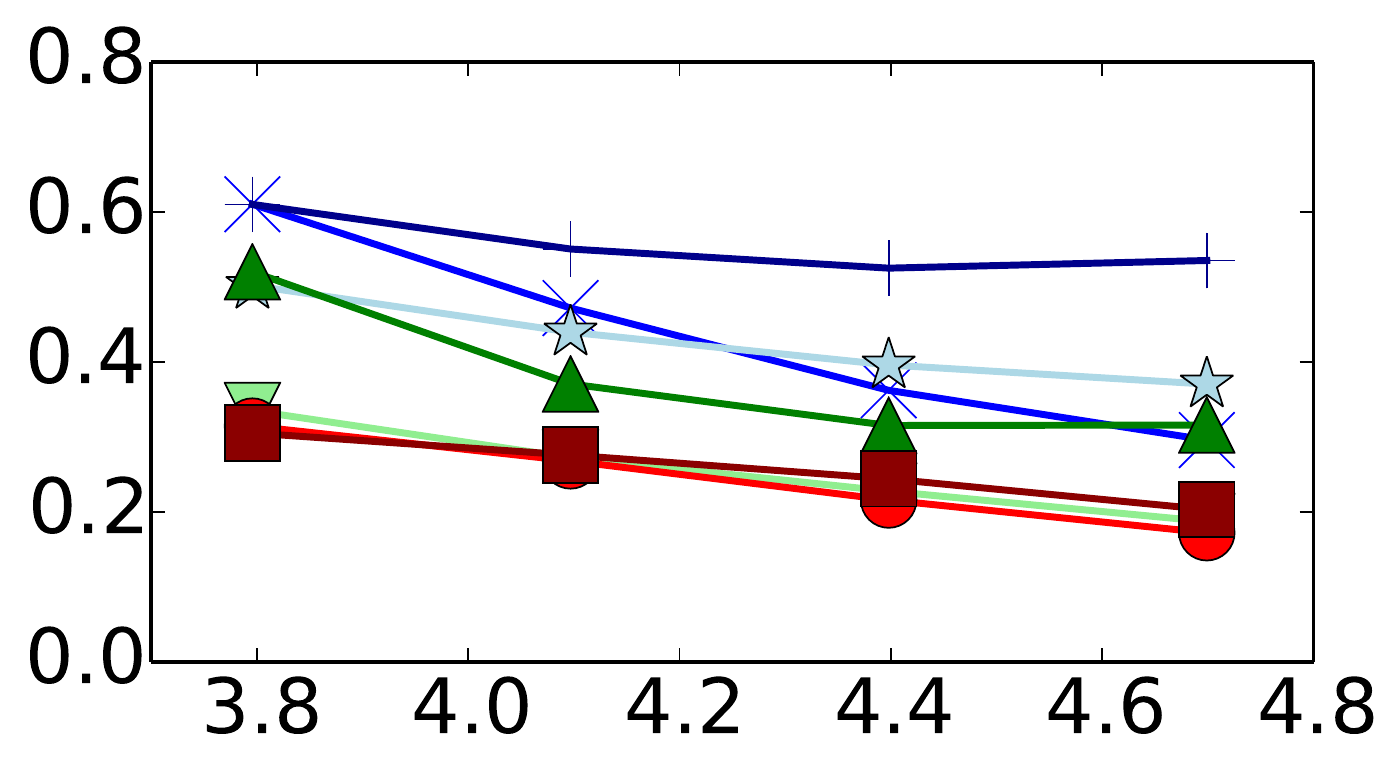}};
\node[inner sep=0pt](cifar10-lenet-mnll) at (5.3, 11.5 + \offset){\includegraphics[width=3.2cm]{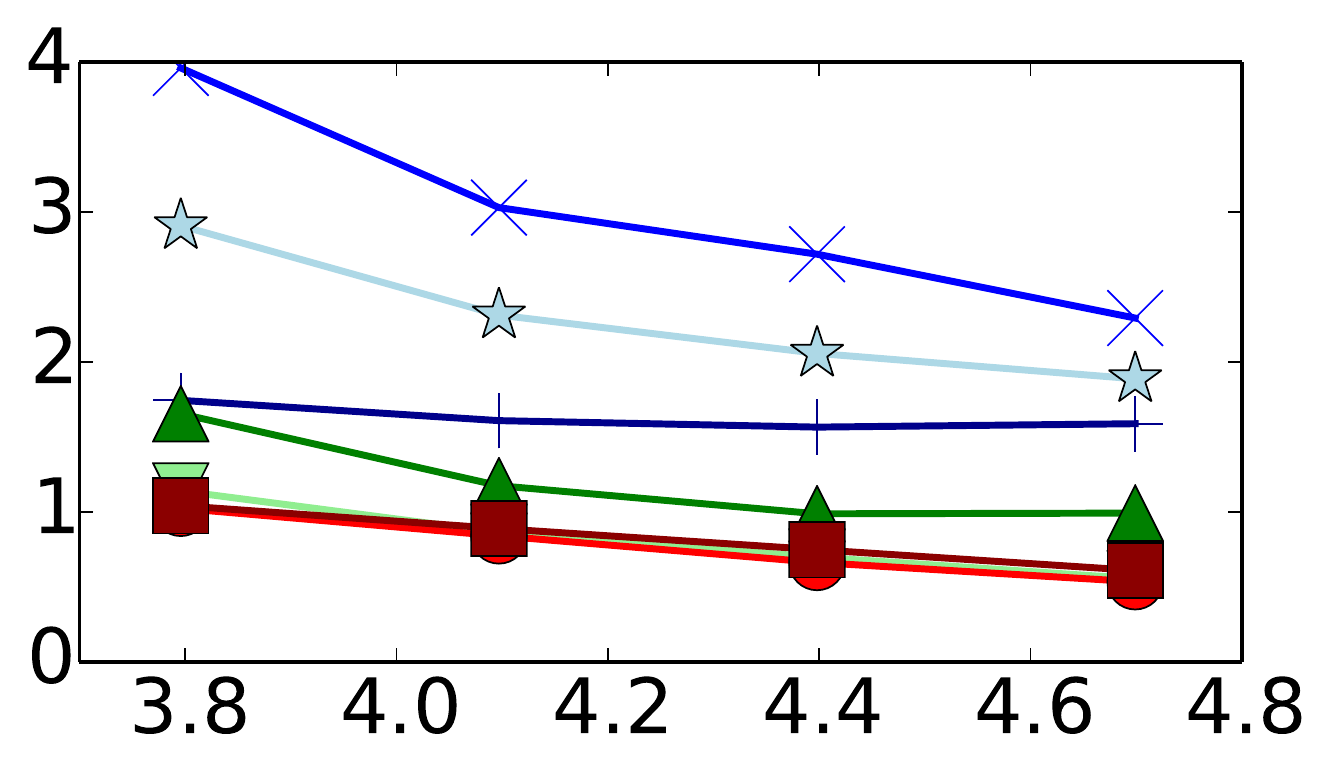}};
\node[inner sep=0pt](cifar10-lenet-ece) at (8.6, 11.5 + \offset){\includegraphics[width=3.2cm]{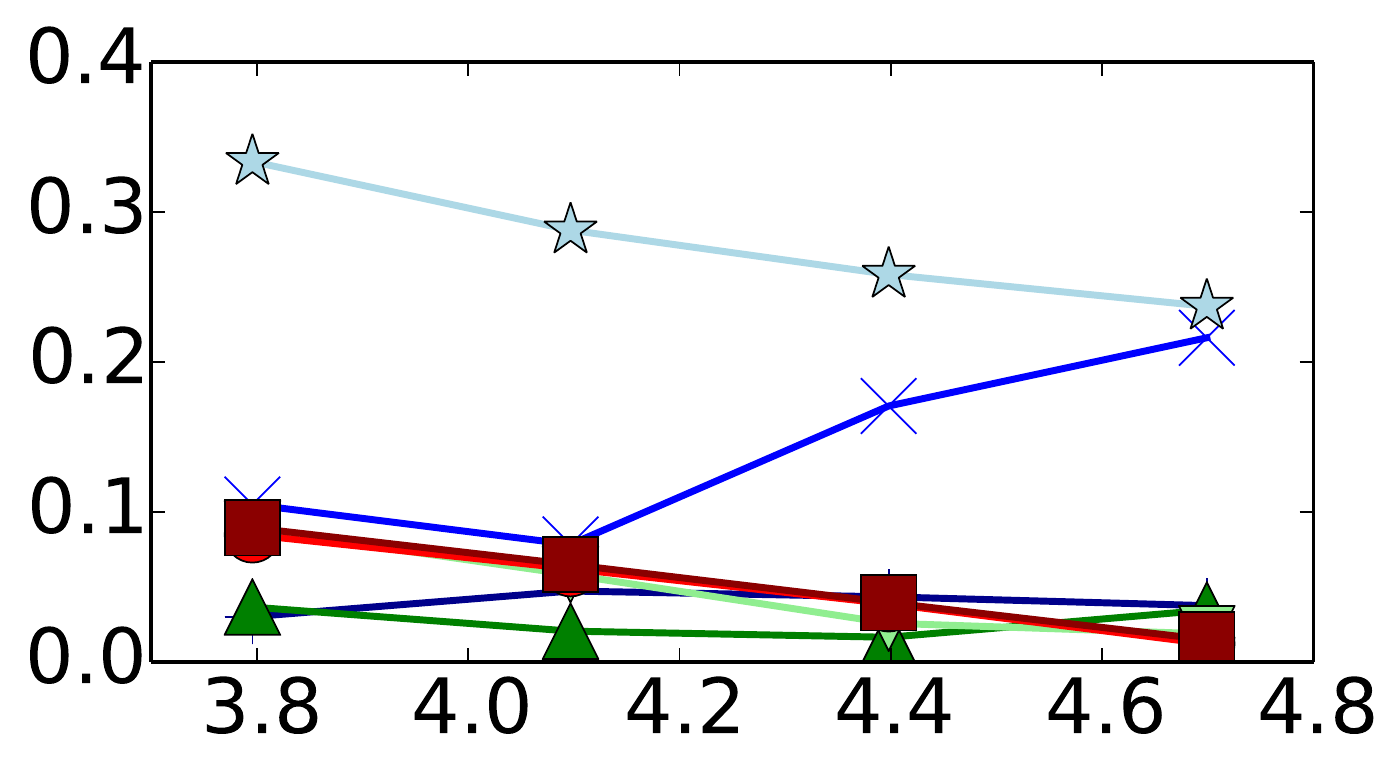}};
\node[inner sep=0pt](cifar10-lenet-brier) at (11.9, 11.5 + \offset){\includegraphics[width=3.2cm]{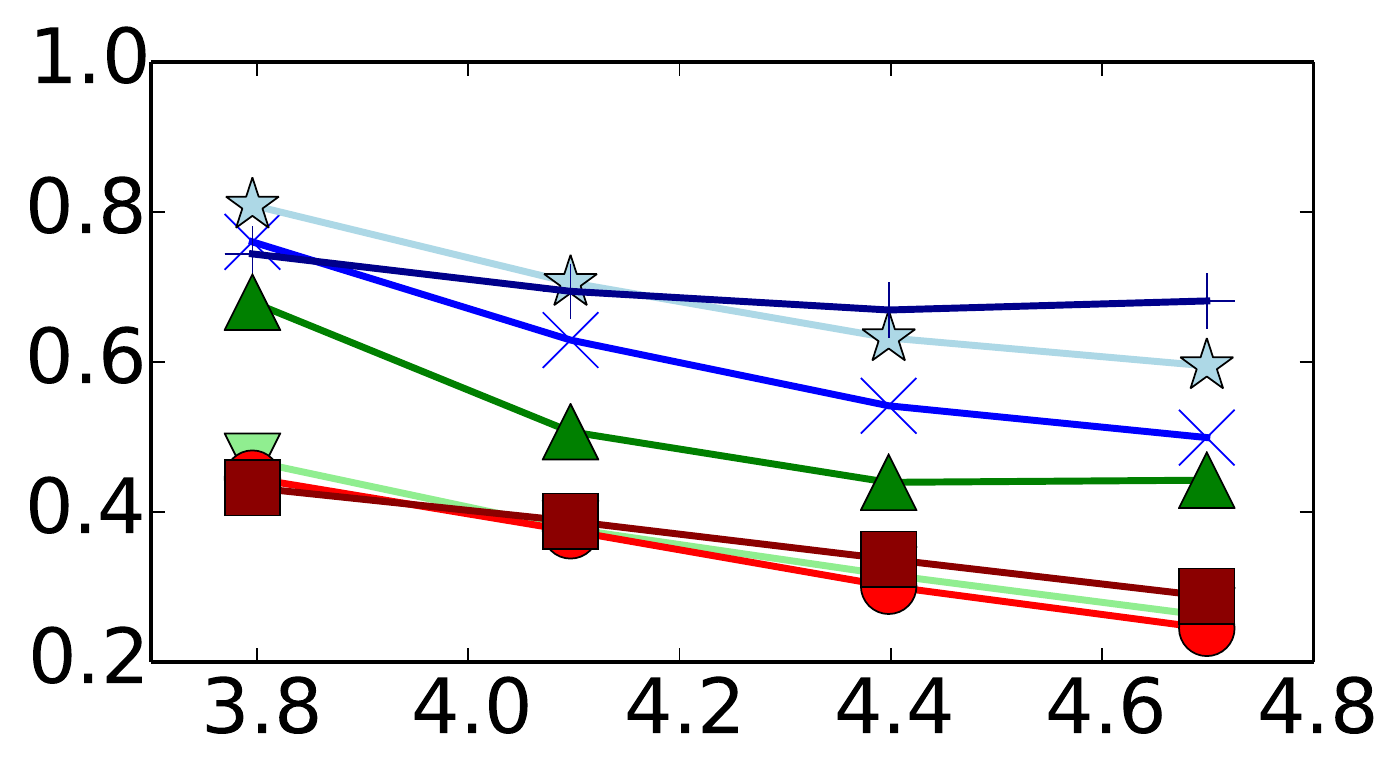}};



\node(mnist-text) at (7, 14.8) {\small DEEP CONVOLUTIONAL STRUCTURE};

\node(mnist-err-text) at (2.0, 14.4) {\err};
\node(mnist-mnll-text) at (5.3, 14.4) {\mnll};
\node(cifar10-err-text) at (8.6, 14.4) {\ece};
\node(cifar10-mnll-text) at (11.9, 14.4) {\brier};

\node[rotate=90](text) at (0.1, 13.3) {\cifart};
\node[rotate=90](text) at (0.1, 11.5) {\cifarh};

\node[inner sep=0pt](cifar10-resnet-err) at (2.0, 13.3){\includegraphics[width=3.2cm]{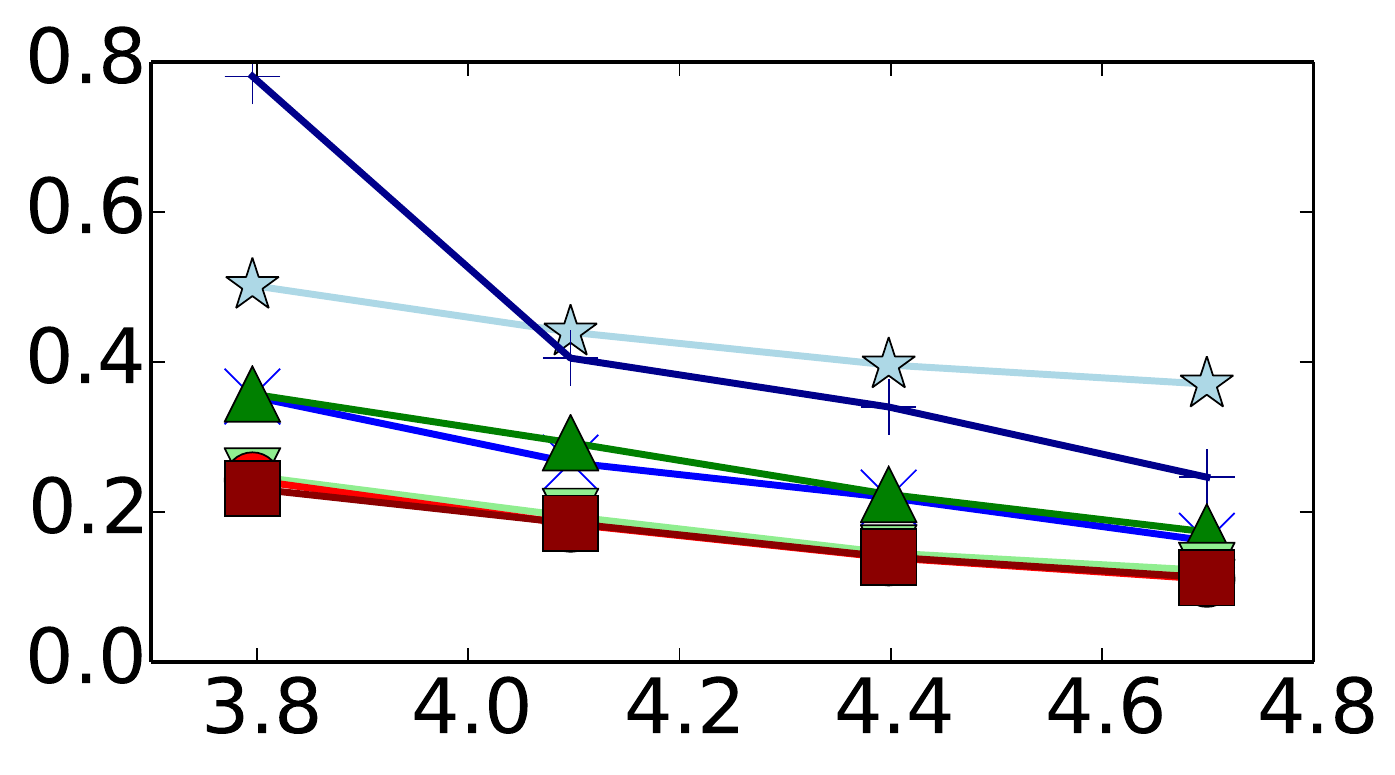}};
\node[inner sep=0pt](mnist-lenet-mnll) at (5.3, 13.3){\includegraphics[width=3.2cm]{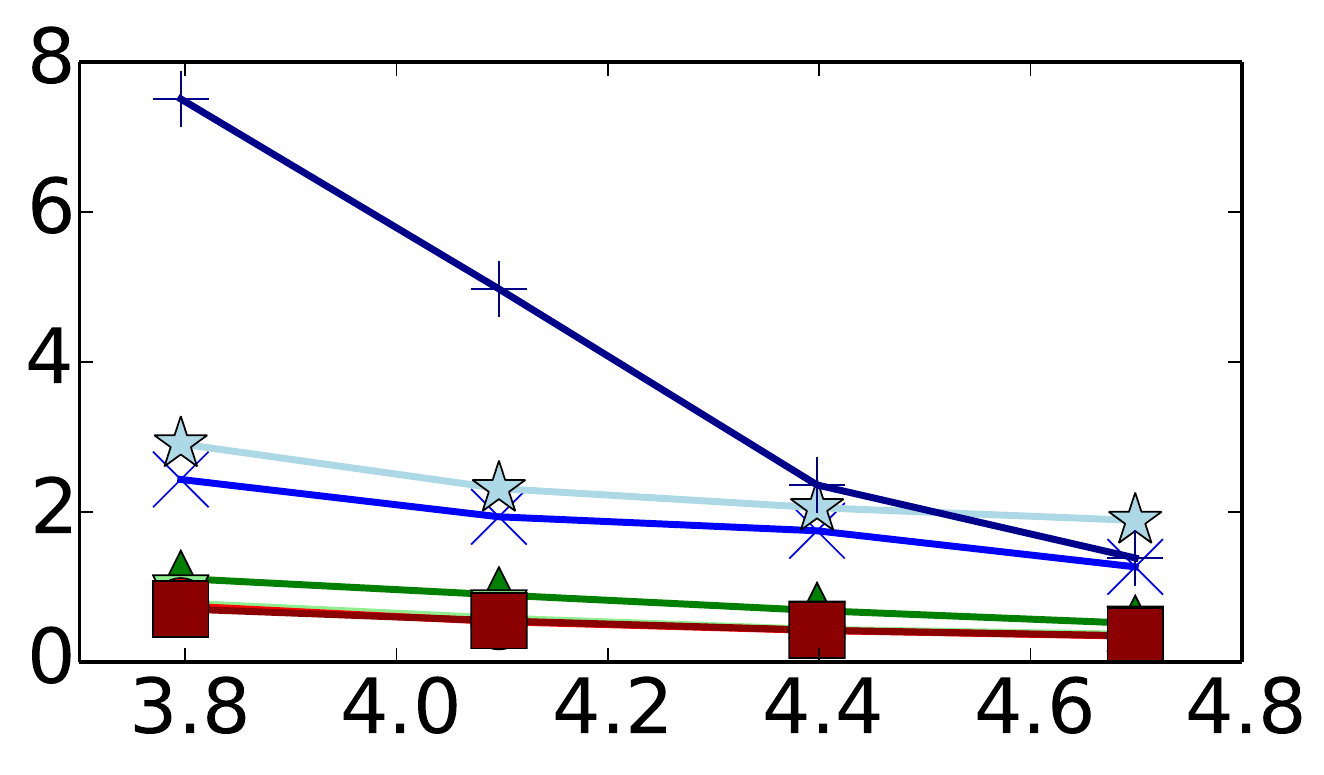}};
\node[inner sep=0pt](mnist-lenet-ece) at (8.6, 13.3){\includegraphics[width=3.2cm]{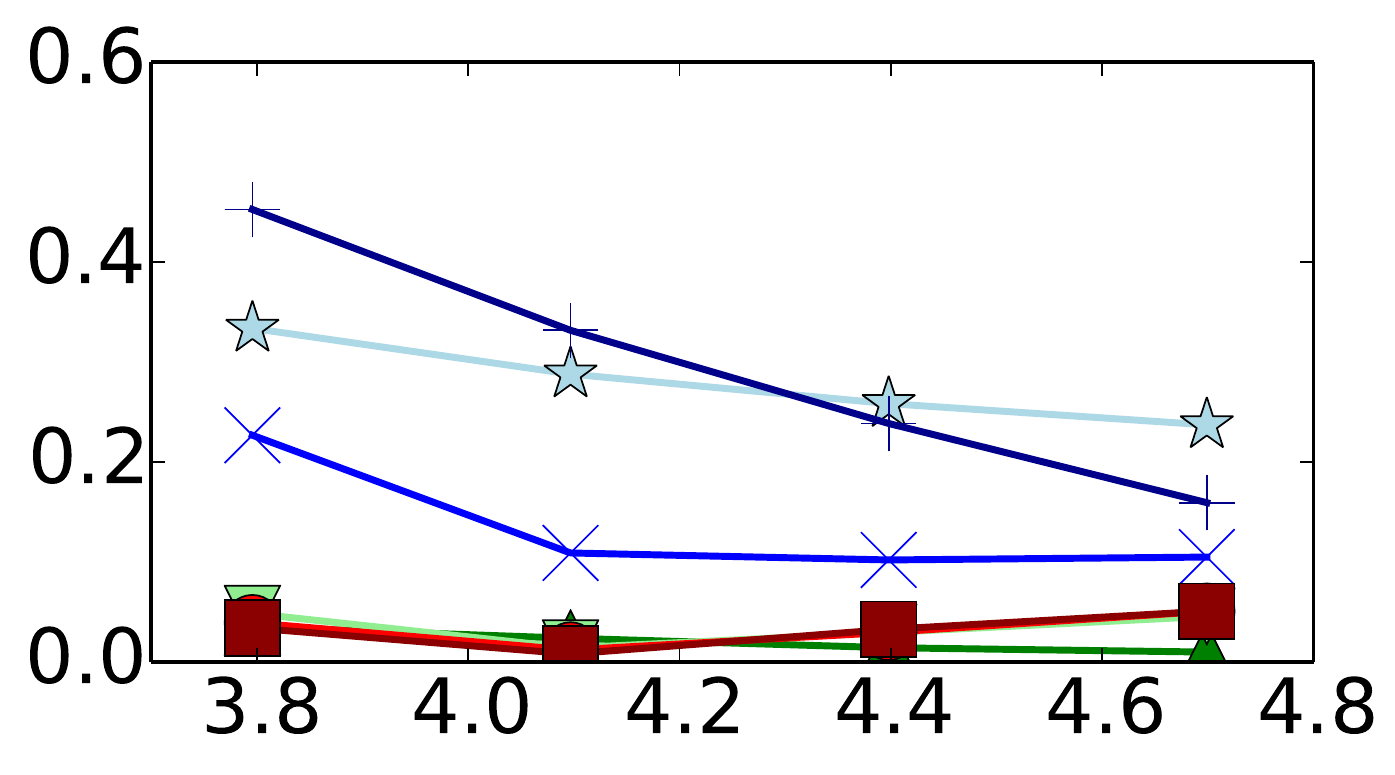}};
\node[inner sep=0pt](cifar10-resnet-brier) at (11.9, 13.3){\includegraphics[width=3.2cm]{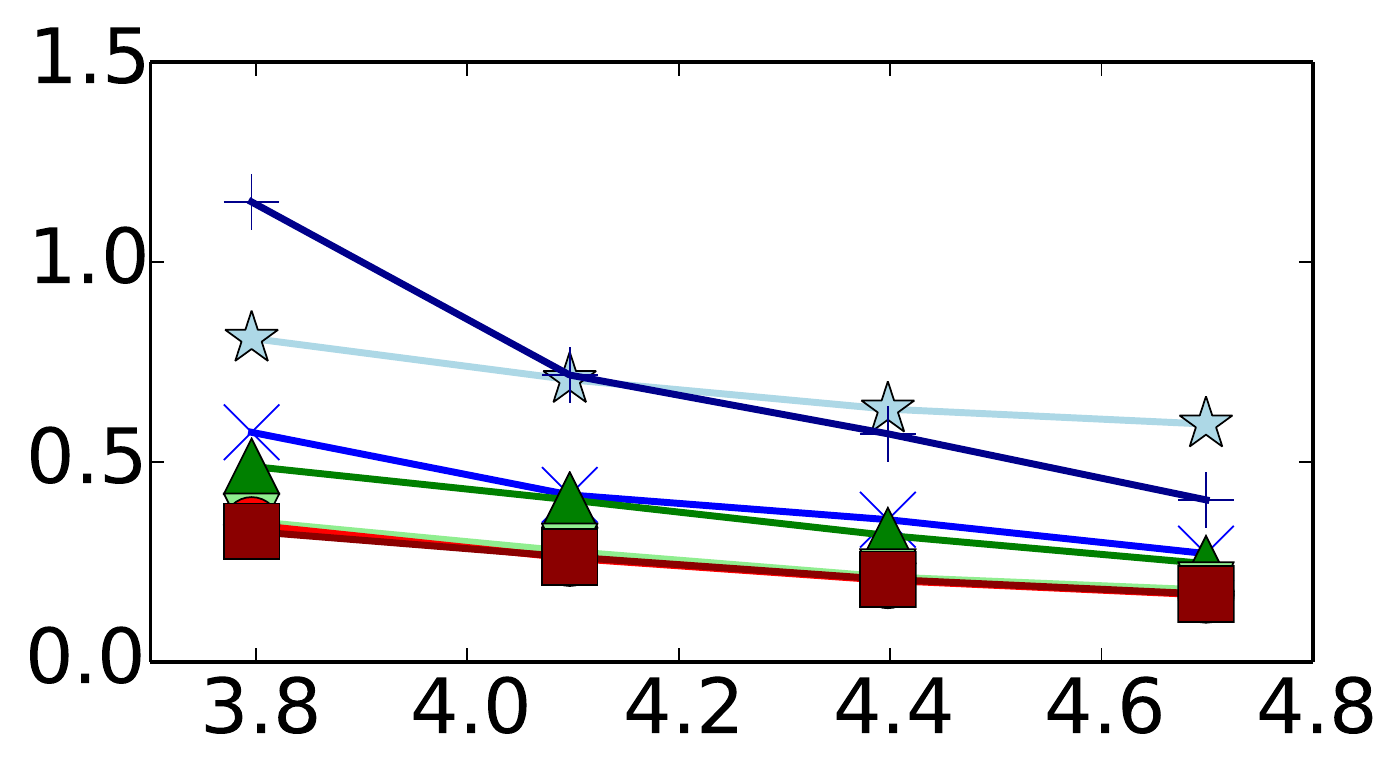}};

\node[inner sep=0pt](cifar100-resnet-err) at (2.0, 11.5){\includegraphics[width=3.2cm]{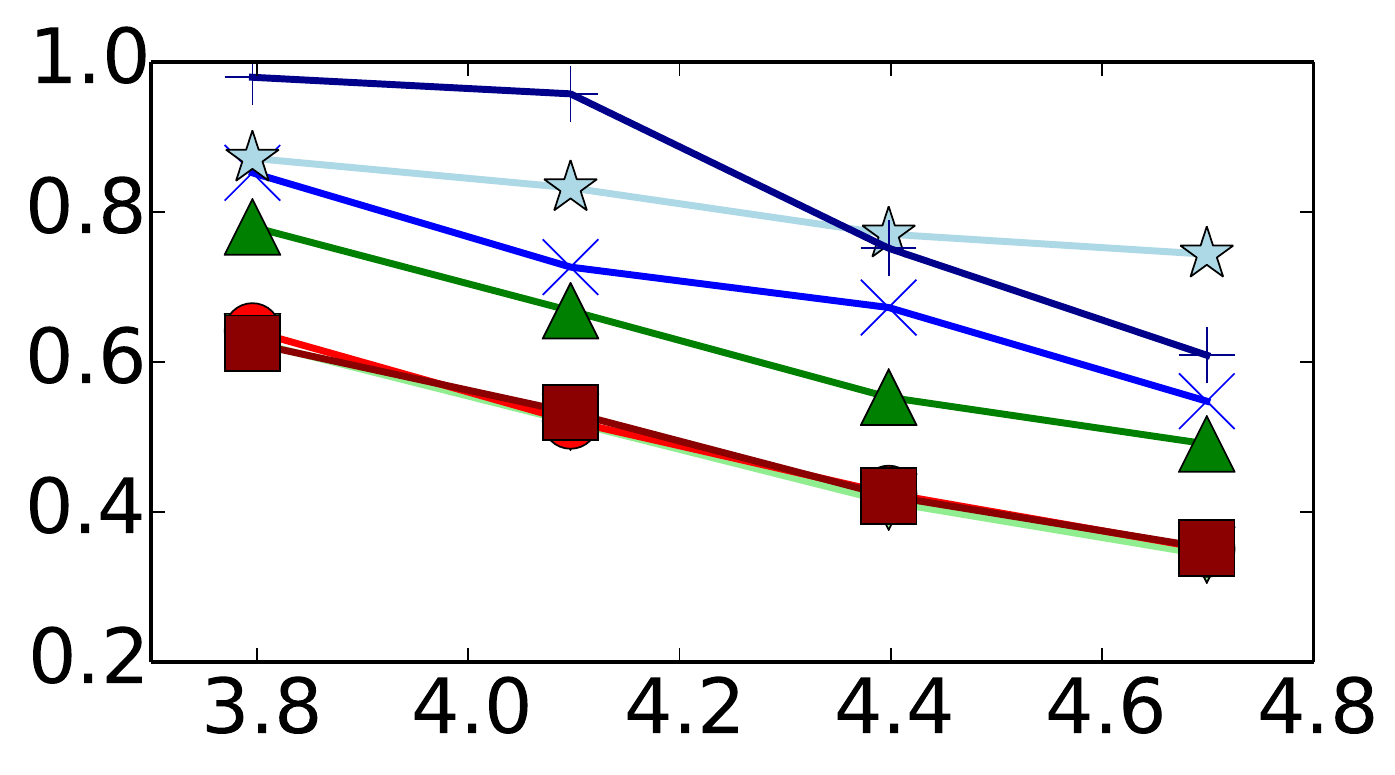}};
\node[inner sep=0pt](cifar100-resnet-mnll) at (5.3, 11.5){\includegraphics[width=3.2cm]{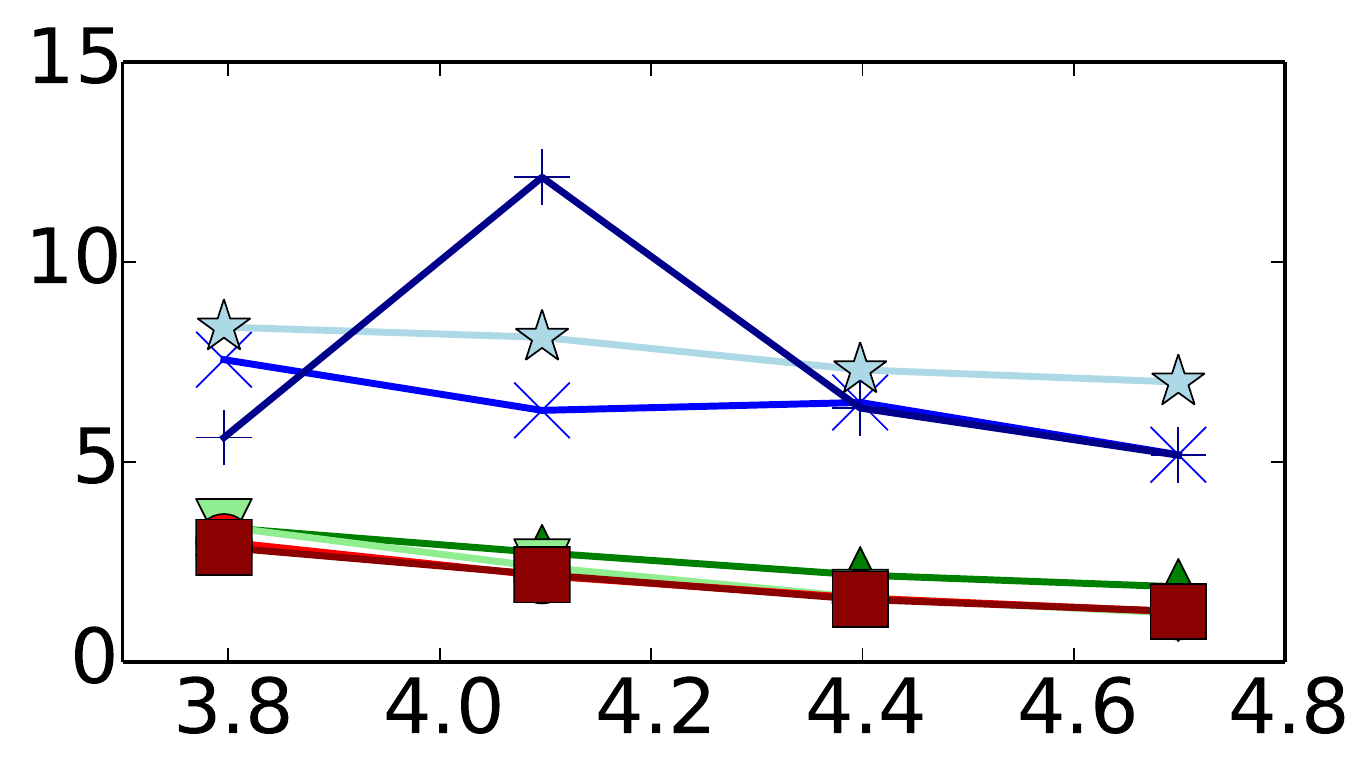}};
\node[inner sep=0pt](cifar100-resnet-ece) at (8.6, 11.5){\includegraphics[width=3.2cm]{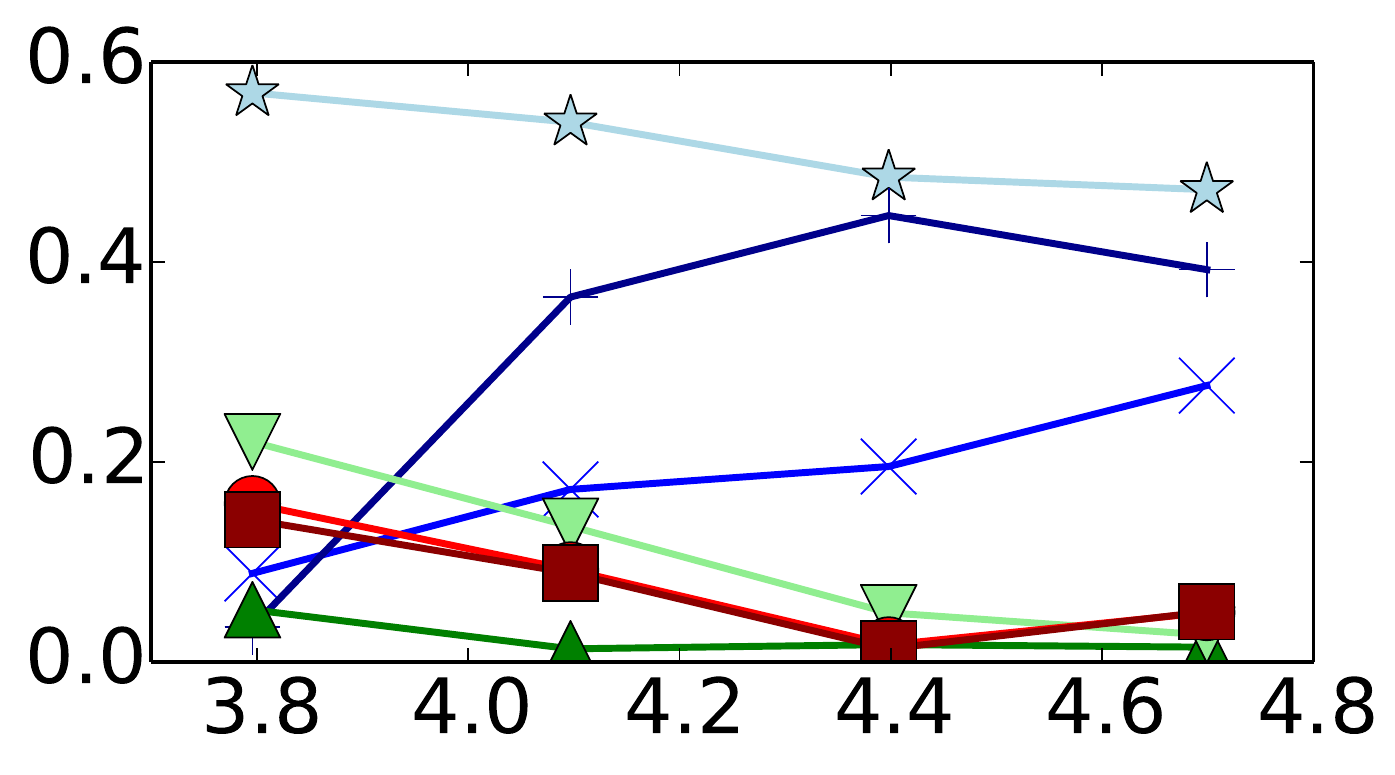}};
\node[inner sep=0pt](cifar100-resnet-brier) at (11.9, 11.5){\includegraphics[width=3.2cm]{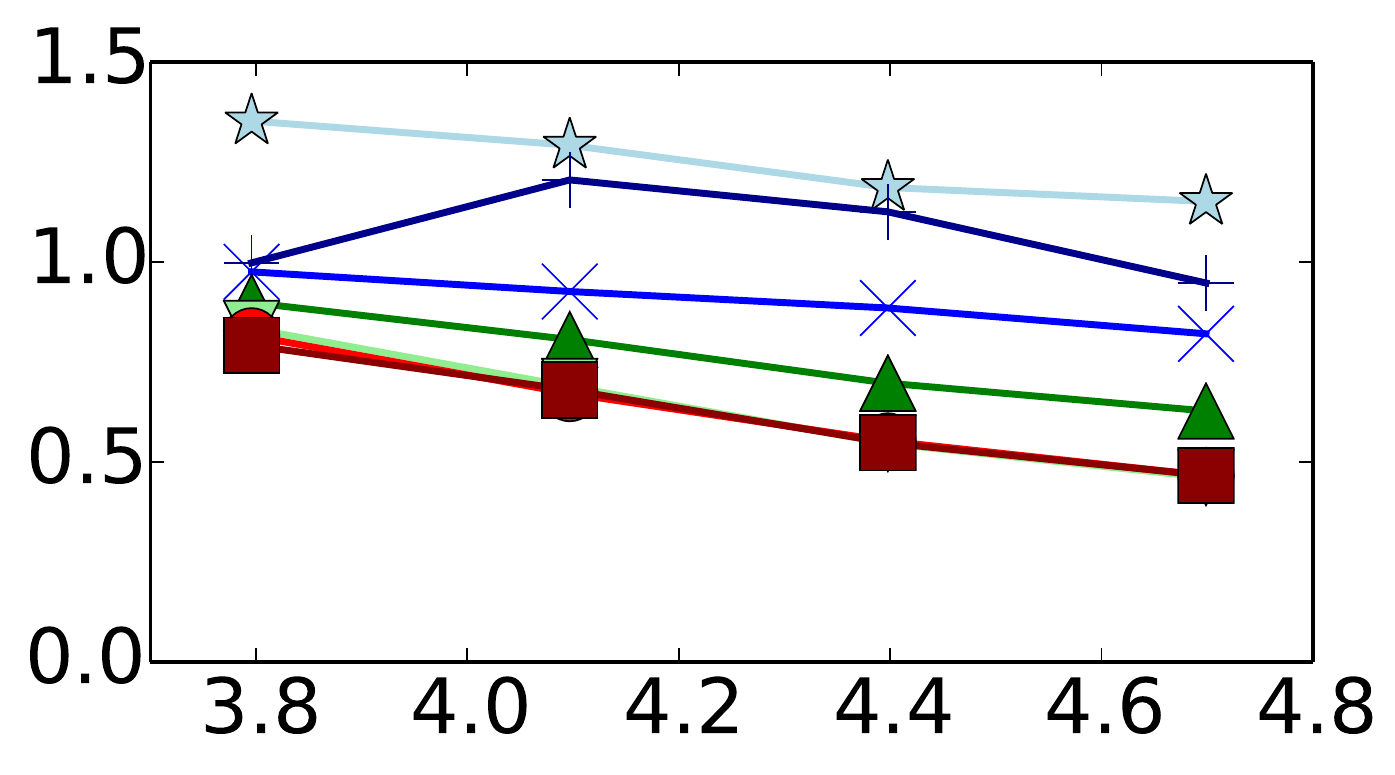}};

\node(iteration1-1) at (2.0, 10.5) {\small $\log_{10}\left(\text{train size}\right)$};
\node(iteration1-1) at (5.3, 10.5) {\small $\log_{10}\left(\text{train size}\right)$};
\node(iteration1-1) at (8.6, 10.5) {\small $\log_{10}\left(\text{train size}\right)$};
\node(iteration1-1) at (11.9, 10.5) {\small $\log_{10}\left(\text{train size}\right)$};









\draw [fill=red](0.9, 9.9) circle (0.10);
\node(text) at (2.1, 9.9) {\cnngprf};

\node [black!40!red] at (0.9, 9.5){$\blacksquare$};
\node(text) at (2.3, 9.5) {\cnngpsorf};

\node [green] at (4.6, 9.9){$\blacktriangledown$};
\node(text) at (5.9, 9.9) {\name{cnn+mcd} \cite{Gal16b}};

\node [black!40!green] at (4.6, 9.5){$\blacktriangle$};
\node(text) at (5.9, 9.5) {\name{cnn+cal}\cite{Guo17}};

\node [blue] at (8.1, 9.9){$\times$};
\node(text) at (9.1, 9.9) {\name{\gpdnn} \cite{Bradshaw17}};

\node [cyan] at (8.1, 9.5){$\star$};
\node(text) at (9.0, 9.5) {\cgp \cite{VanDerWilk17}};

\node [black!40!blue] at (11.1, 9.9){$+$};
\node(text) at (12.2, 9.9) {\svdkl\cite{Wilson16}};






\end{tikzpicture}
\caption{Comparison of our \cnngprf and \cnngpsorf with existing combinations of \cnns with \gps, and with Bayesian \cnns and post-calibrated \cnns.}
\label{model comparison}
\end{center}
\vskip -0.2in
\end{figure}

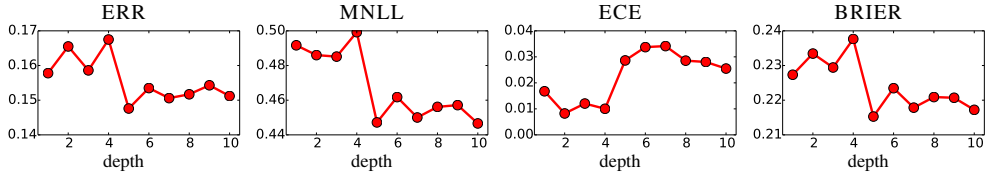
\begin{figure}[t]
\vskip 0.2in
\begin{center}
\begin{tikzpicture}
\node(mnist-err-text) at (2.0, 14.0) {\err};
\node(mnist-mnll-text) at (5.3, 14.0) {\mnll};
\node(cifar10-err-text) at (8.6, 14.0) {\ece};
\node(cifar10-mnll-text) at (11.9, 14.0) {\brier};


\node[inner sep=0pt](mnist-lenet-err) at (2.0, 13){\includegraphics[width=3.2cm]{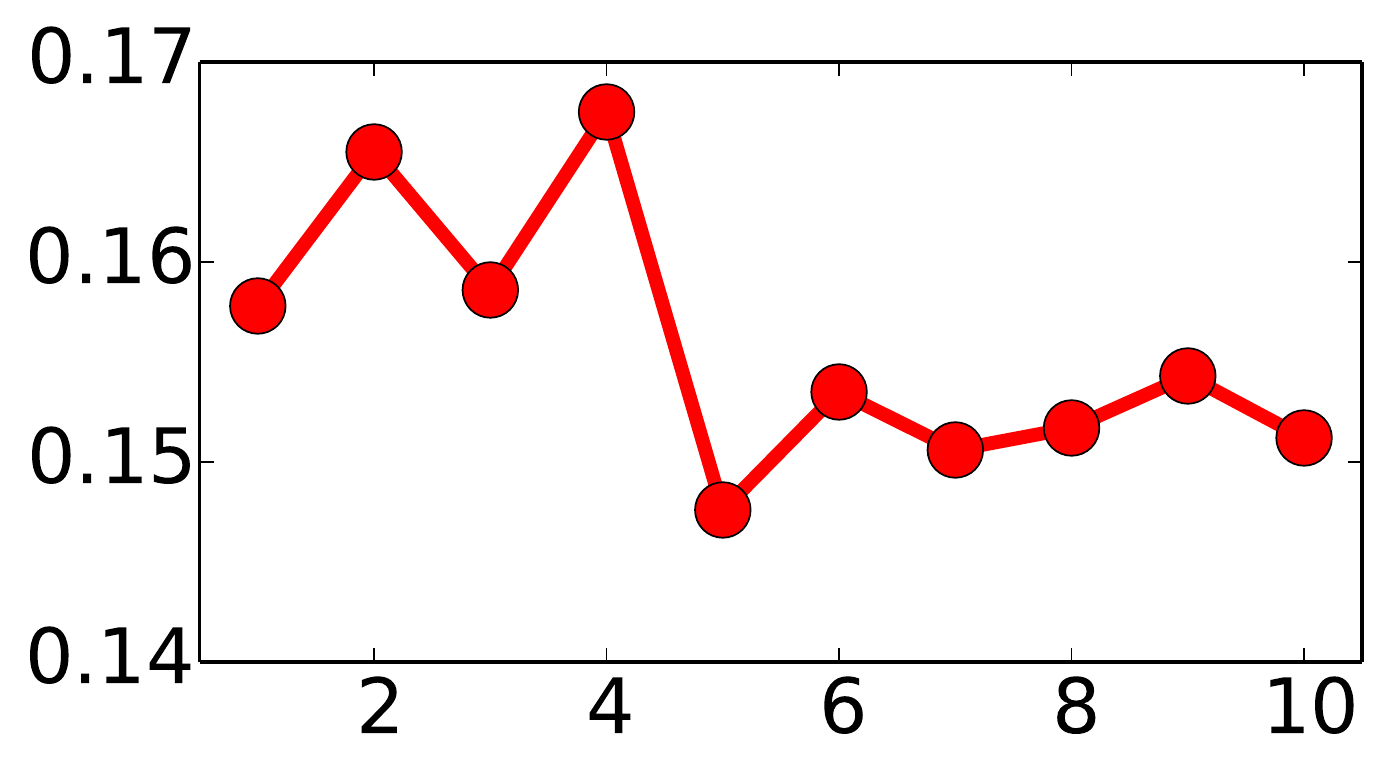}};
\node[inner sep=0pt](mnist-lenet-mnll) at (5.3, 13){\includegraphics[width=3.2cm]{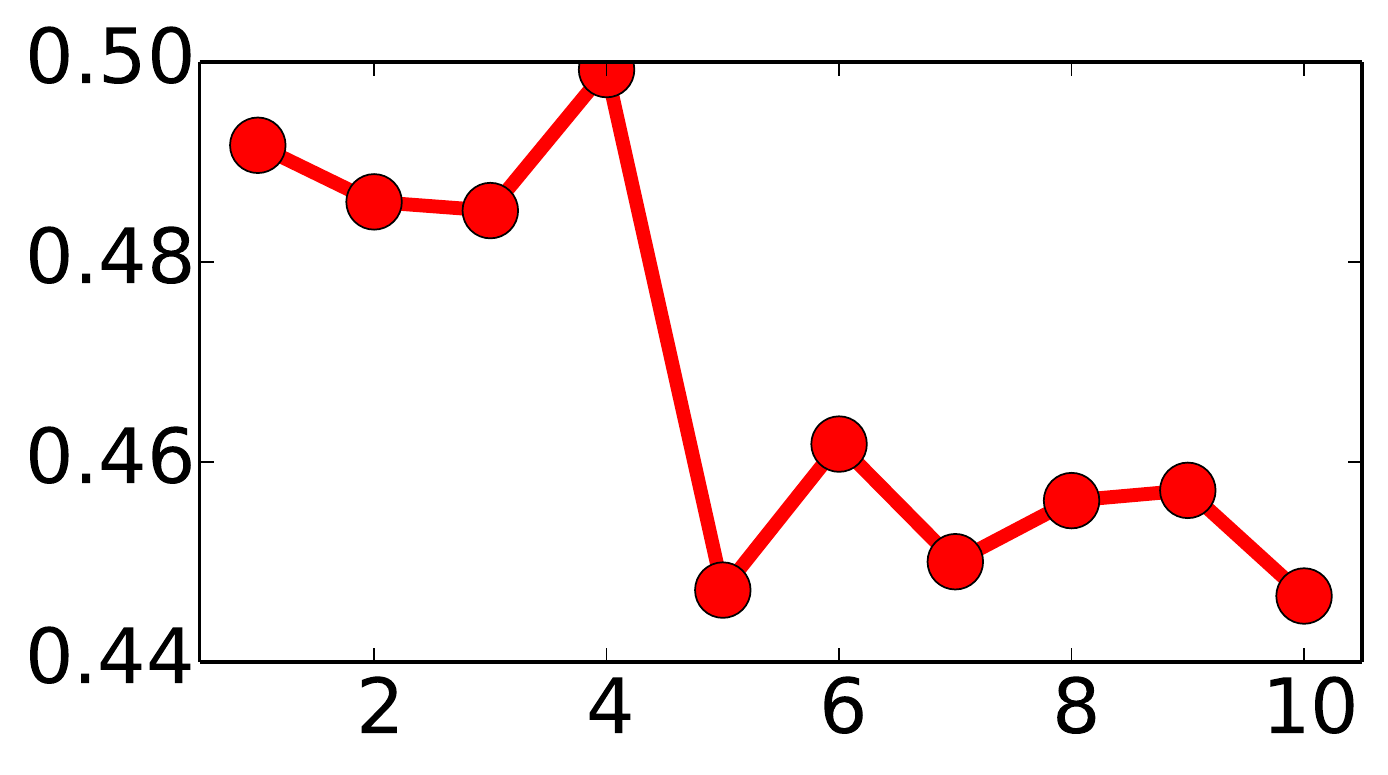}};
\node[inner sep=0pt](mnist-lenet-ece) at (8.6, 13){\includegraphics[width=3.2cm]{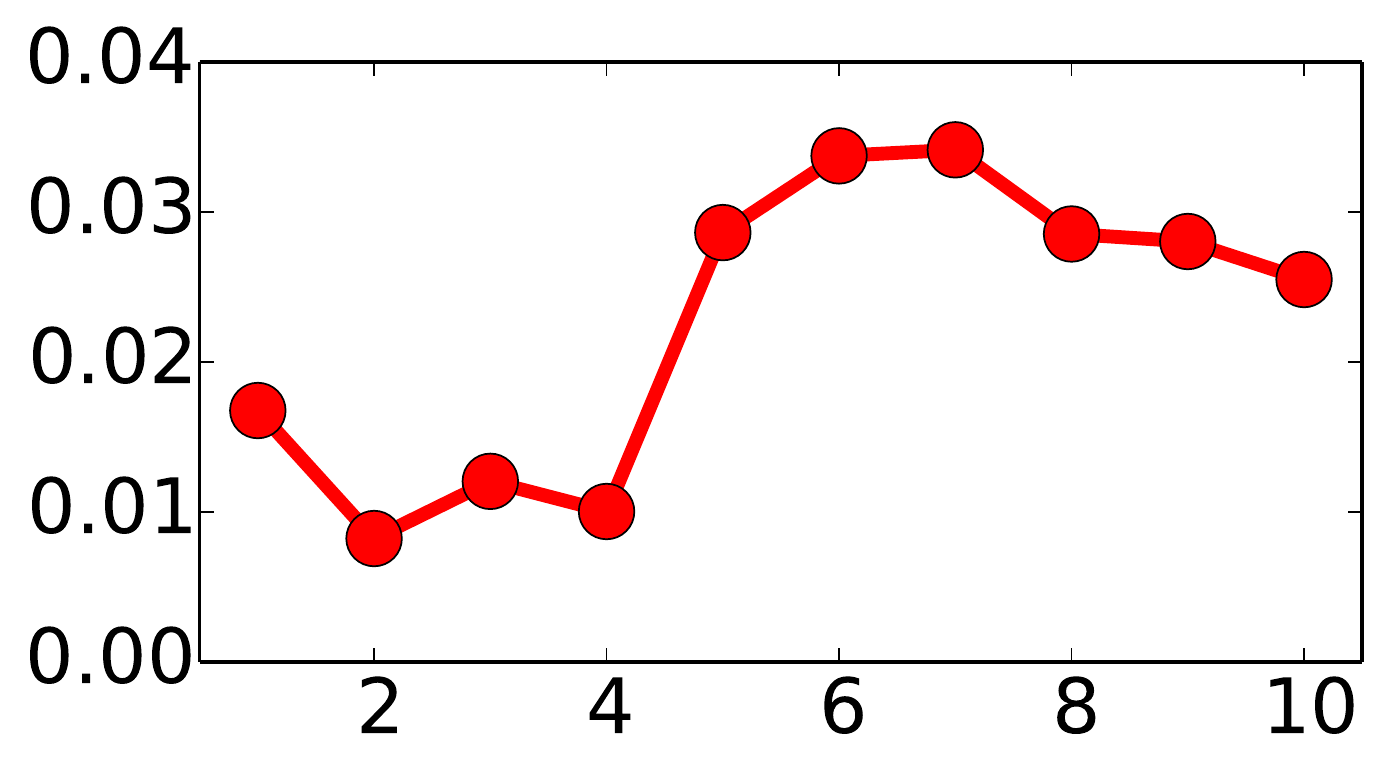}};
\node[inner sep=0pt](mnist-lenet-brier) at (11.9, 13){\includegraphics[width=3.2cm]{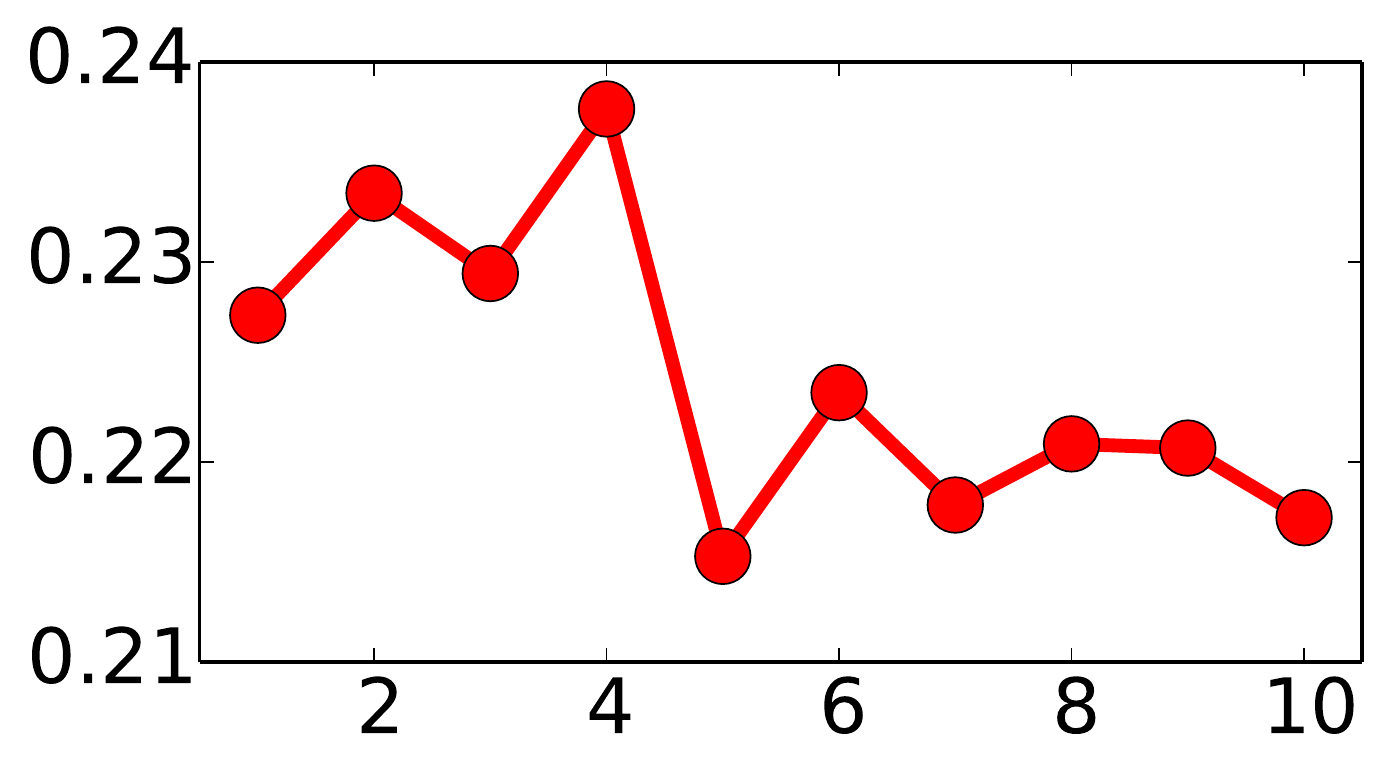}};


\node(iteration1-1) at (2.0, 12.0) {\scriptsize depth};
\node(iteration1-1) at (5.3, 12.0) {\scriptsize depth};
\node(iteration1-1) at (8.6, 12.0) {\scriptsize depth};
\node(iteration1-1) at (11.9, 12.0) {\scriptsize depth};





\end{tikzpicture}
\caption{Performance of the proposed model when varying the depth of the \dgp on top of a \resnet convolutional structure on \cifart dataset.}
\label{depth comparison}
\end{center}
\vskip -0.2in
\end{figure}

\paragraph{Experiments combining \cnns and Deep \gps:}
In figure~\ref{depth comparison}, we report results varying the depth of a \dgp on top of the convolutional layers; again, we learn the convolutional filters and the \dgp end-to-end as discussed in the previous sections.
We show results when applying our model to the whole \cifart data set in the case of the shallow convolutional structure (table~\ref{tab:configurations:cnns}).
We feed-forward the convolutional features to all layers of the \dgp, in line with what suggested in the literature of \dgps to avoid pathologies in the functions that can be modeled \cite{Cutajar17,Duvenaud14,Neal96}. 
The results indicate that increasing the complexity of the model improves on all performance metrics, and worsen calibration, which however is still around $3\%$ \ece.
This is in line with the intuition that increasing model complexity negatively impacts calibration.

\paragraph{Knowing when the model doesn't know:}

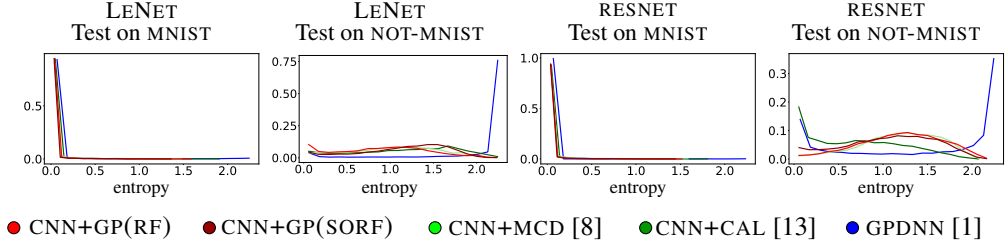
\begin{figure}[t]
\vskip 0.2in
\begin{center}
\begin{tikzpicture}
\node(text) at (2.0, 14.4)    {\small \lenet};
\node(text) at (2.0, 14.1)    {\small Test on \mnist};
\node(text) at (5.3, 14.4)   {\small \lenet};
\node(text) at (5.3, 14.1)   {\small Test on \notmnist};
\node(text) at (8.6, 14.4)  {\small \resnet};
\node(text) at (8.6, 14.1)    {\small Test on \mnist};
\node(text) at (11.9, 14.4){\small \resnet};
\node(text) at (11.9, 14.1)   {\small Test on \notmnist};


\node[inner sep=0pt](mnist-lenet-err) at (2.0, 13){\includegraphics[width=3.2cm]{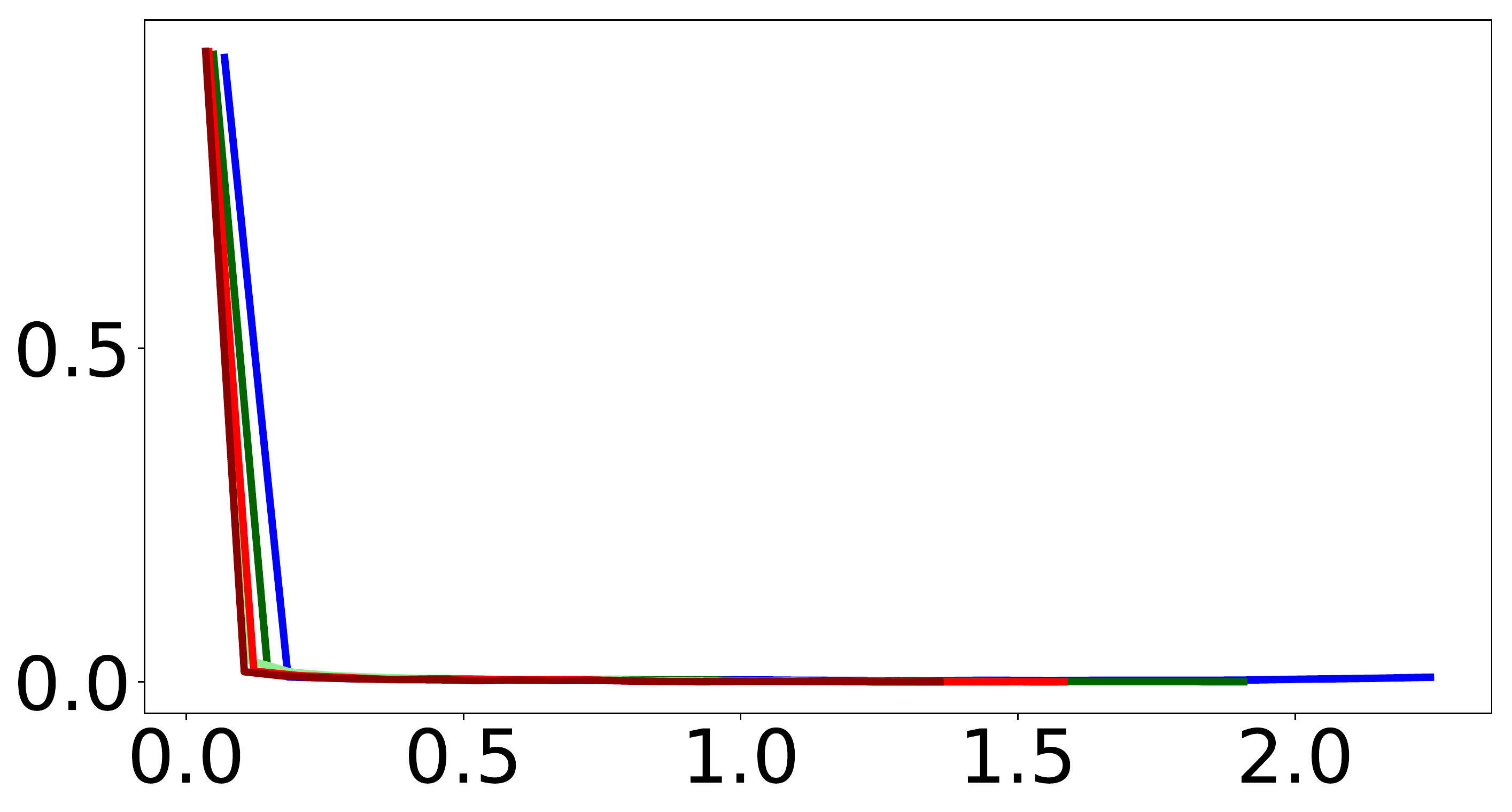}};
\node[inner sep=0pt](mnist-lenet-mnll) at (5.3, 13){\includegraphics[width=3.2cm]{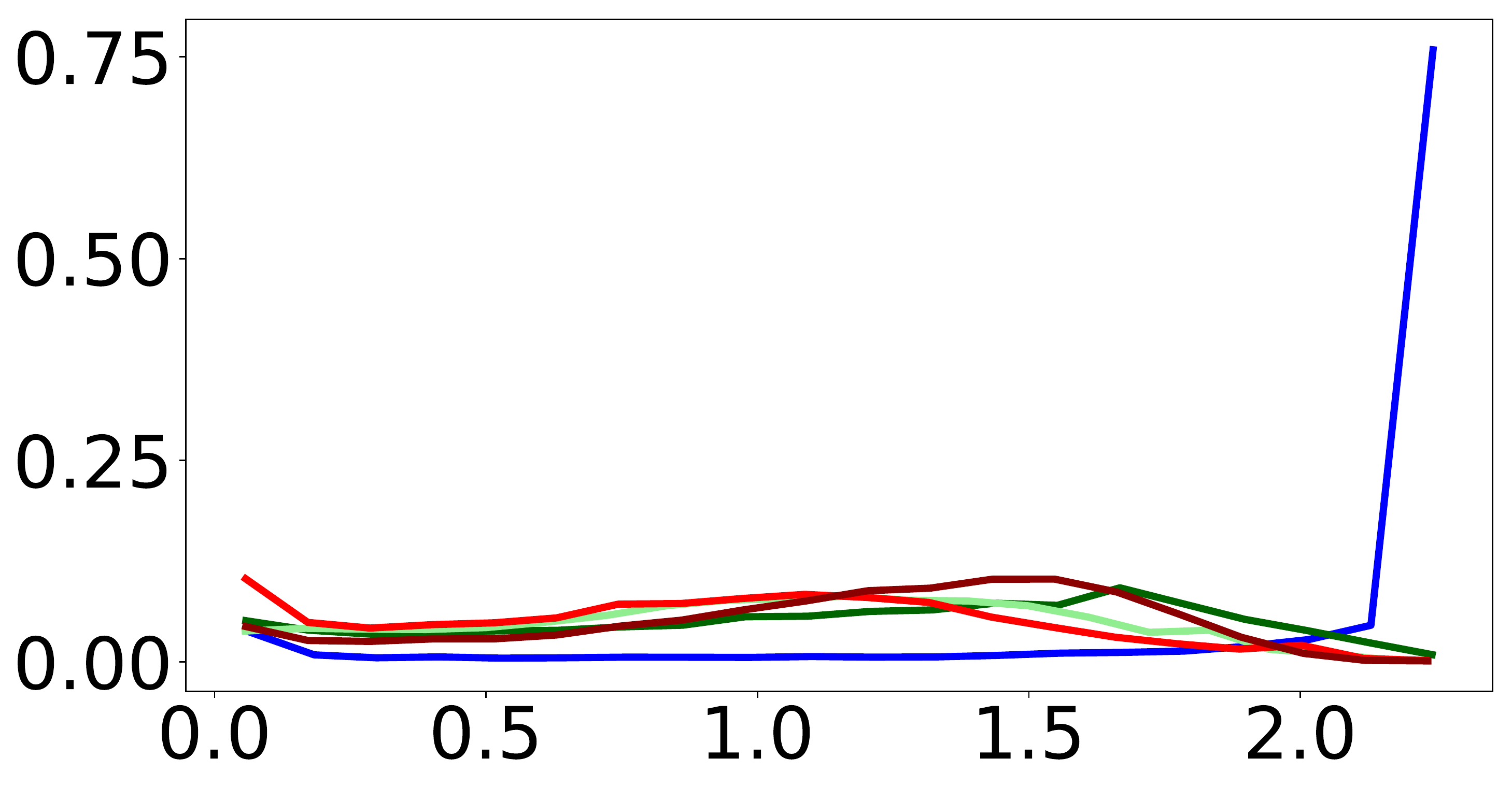}};
\node[inner sep=0pt](mnist-lenet-ece) at (8.6, 13){\includegraphics[width=3.2cm]{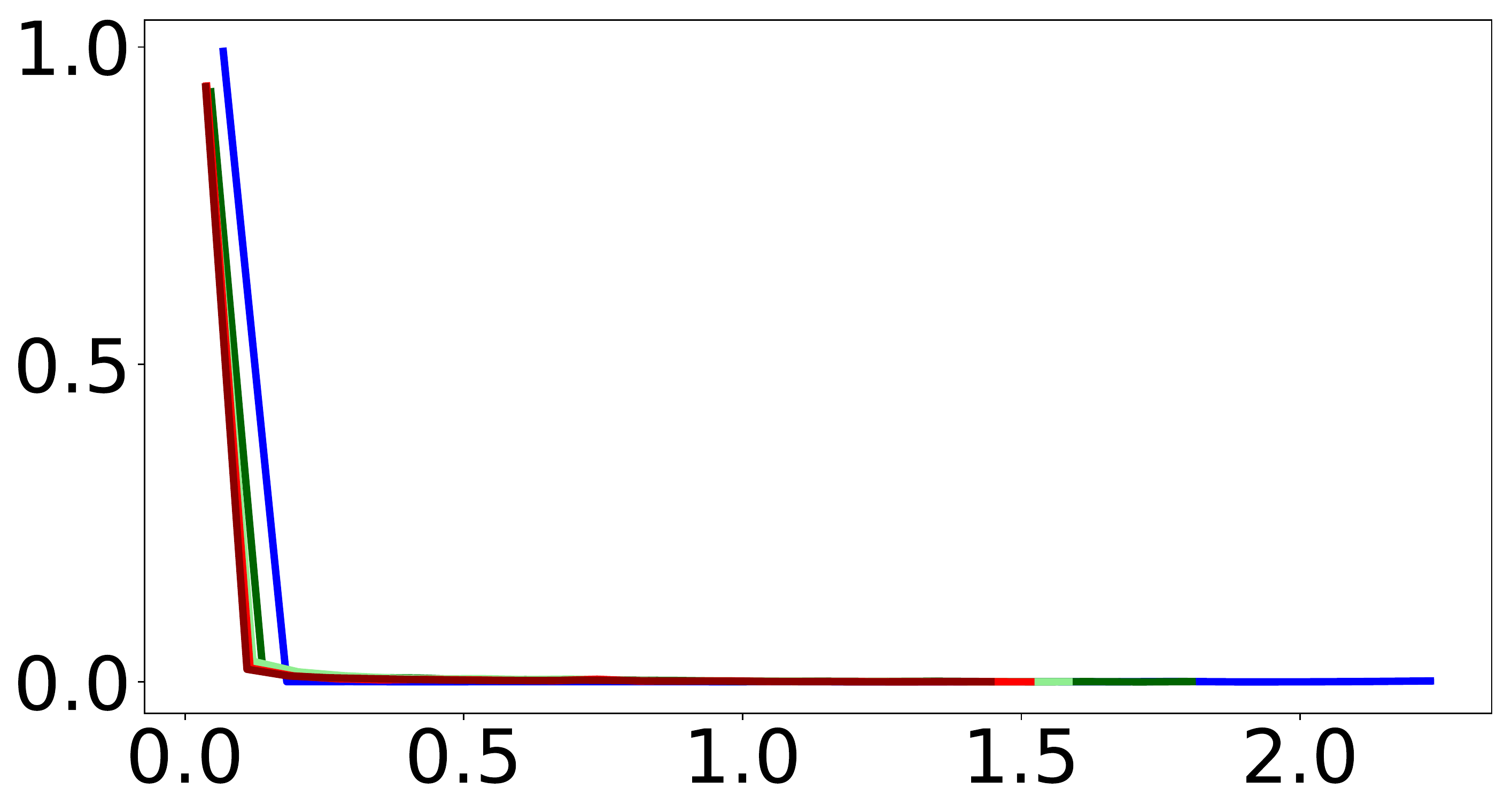}};
\node[inner sep=0pt](mnist-lenet-brier) at (11.9, 13){\includegraphics[width=3.2cm]{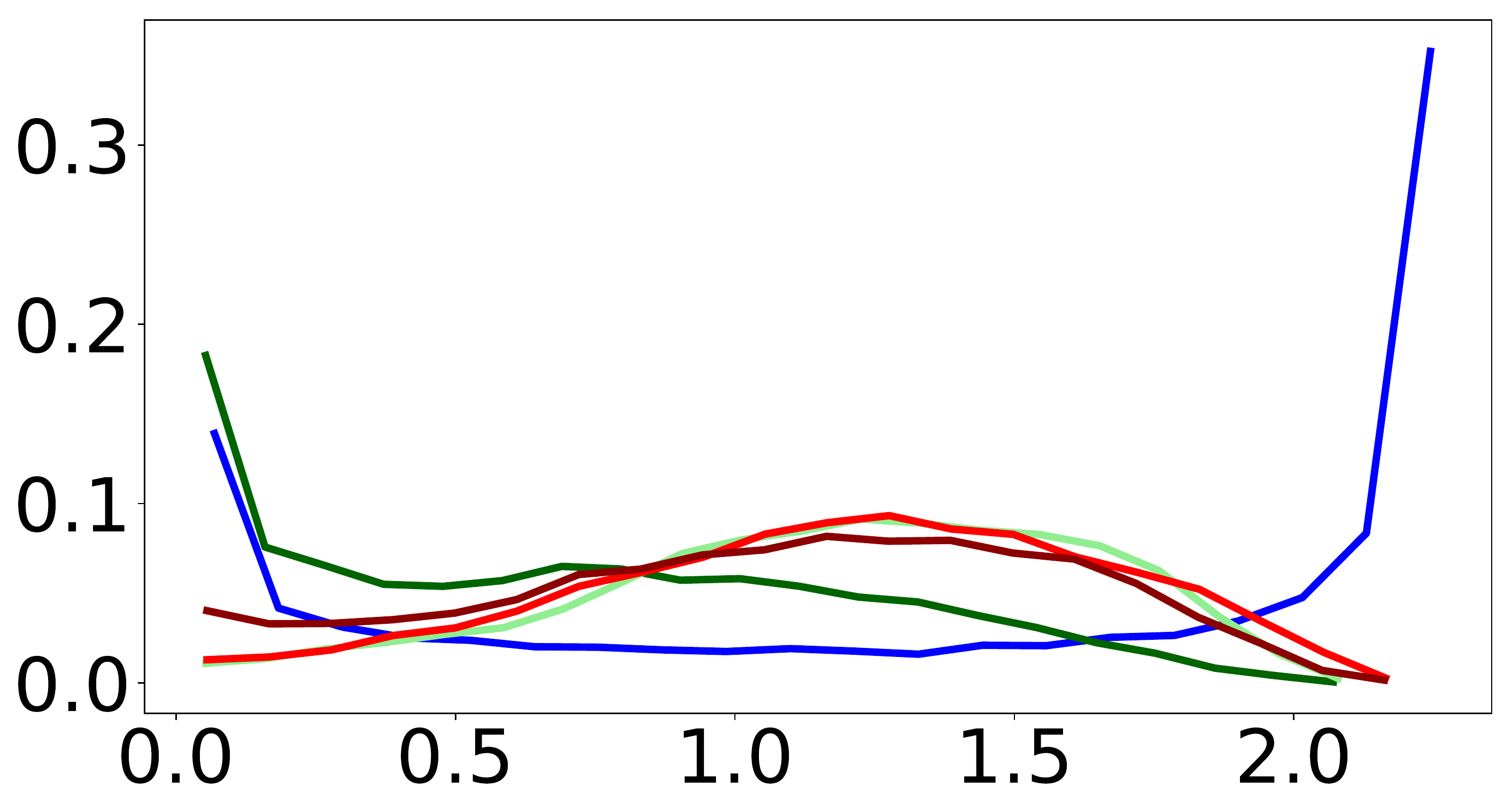}};


\node(iteration1-1) at (2.0, 12.0) {\scriptsize entropy};
\node(iteration1-1) at (5.3, 12.0) {\scriptsize entropy};
\node(iteration1-1) at (8.6, 12.0) {\scriptsize entropy};
\node(iteration1-1) at (11.9, 12.0) {\scriptsize entropy};

\draw [fill=red](0.3, 11.5) circle (0.08);
\node(text) at (1.45, 11.5) {\cnngprf};

\draw [fill=black!40!red](2.9, 11.5) circle (0.08);
\node(text) at (4.2, 11.5) {\cnngpsorf};

\draw [fill=green](5.9, 11.5) circle (0.08);
\node(text) at (7.1, 11.5) {\name{cnn+mcd} \cite{Gal16b}};

\draw [fill=black!40!green](8.7, 11.5) circle (0.08);
\node(text) at (9.9, 11.5) {\name{cnn+cal} \cite{Guo17}};

\draw [fill=blue](11.45, 11.5) circle (0.08);
\node(text) at (12.4, 11.5) {\name{\gpdnn} \cite{Bradshaw17}};

\end{tikzpicture}
\caption{Density plot of predictive entropies when the models trained on \mnist are tested on \mnist and \notmnist. We report results for two different depths of the convolutional structure. \notmnist dataset available at \url{http://yaroslavvb.blogspot.fr/2011/09/notmnist-dataset.html}}
\label{entropy predictions}
\end{center}
\vskip -0.2in
\end{figure}

We report experiments showing the ability of our model to know when it does not know, following a similar experimental setup as in \cite{Lakshminarayanan17}.
In this experiment we train our \cnngprf model on \mnist and test on the \notmnist dataset, which contains images of letters from ``A'' to ``J'' in various typefaces.
For this experiment, while we do not know the exact value that we should obtain for predictive probabilities, we expect to observe low entropy in the predictions when tesing on \mnist and high entropy when predicting on \notmnist, indicating high uncertainty.
The results are reported in figure~\ref{entropy predictions}, where we show the density plot of the entropy of predictive probabilities for two depths of the convolutional structure. 
In the figure, we compare our \cnngprf against one of the methods combining \cnns and \gps, that is \gpdnn.
In the figure, we also include results on \cnns with post-calibration and Bayesian \cnns inferred with \mcd. 
Our approach is competitive with Bayesian \cnns and it is considerably superior to post-calibration. 
This is especially true in the case of a deeper convolutional structure, where post-calibration still yields a large number of predictions with low uncertainty. 
Interestingly, \gpdnn assigns large uncertainty to predictions on \notmnist, although with the deeper convolutional architecture it yields a large fraction of predictions with low entropy.

\section{Conclusions}

Despite the considerable interest in combining \cnns with \gps, little attention has been devoted to understand the implications in terms of the ability of these models to accurately quantify the level of uncertainty in predictions. 
This is the first work that highlights the issues of calibration of these models, showing that \gps cannot cure the issues of miscalibration in \cnns. 
We have proposed a novel combination of \cnns and \gps where the resulting model becomes  a particular form of a Bayesian \cnn for which inference using variational inference is straightforward. 
%
%
%
However, our results also indicate that combining \cnns and \gps does not significantly improve the performance of standard \cnns. This can serve as a motivation for investigating new approximation methods for scalable inference in \gp models and combinations with \cnns. 


\subsubsection*{Acknowledgments}

JPC acknowledges support from the Simons Foundation and the McKnight Foundation.
MF gratefully acknowledges support from the AXA Research Fund.



\small





\normalsize

\newpage

\appendix

\normalsize

\section{Random Feature Expansion of the \rbf Covariance}

We report here the expansion of the popular Radial Basis Function (\rbf) covariance. 
Following the convolutional representation of images in our \cnngprf model, the \rbf covariance is defined as:
\begin{equation}
	\label{eq:ARD_kernel}
  k_{\mathrm{rbf}}(\mathbf{x}_i, \mathbf{x}_j | \convweights, \thetavect) = \sigma^2 \exp\left[- 
 \left(\cvect(\xvect_i | \convweights) - \cvect(\xvect_j | \convweights)\right)^{\top} \Lambdamat^{-1} \left(\cvect(\xvect_i | \convweights) - \cvect(\xvect_j | \convweights)\right)
\right]   \text{,}
 \end{equation} 
with $\thetavect =  (\sigma, \Lambdamat=\diag(\ell_1^2, \ldots, \ell_\nconv^2))$.

It is possible to express this covariance function as the Fourier transform of a non-negative measure $p(\omegavect)$ \cite{Rahimi08},
where $\omegavect$ are the so-called spectral frequencies. 
%
It is straightforward to verify that $p\left(\omegavect\right) = \norm\left(\omegavect | \zerovect, \Lambdamat^{-1} \right) 
$.
Stacking $N_{\mathrm{RF}}$ Monte Carlo samples from $p(\omegavect)$ into $\Omegamat$ by column, we obtain
\begin{equation}
\Phimat_{\mathrm{rbf}} = \sqrt{\frac{\sigma^2}{N_{\mathrm{RF}}}} \left[ \cos\left(\Cmat(\Xmat | \convweights) \, \Omegamat\right), \sin\left(\Cmat(\Xmat | \convweights) \, \Omegamat\right) \right] \text{,}
\end{equation}
where $\Cmat(\Xmat | \convweights)$ denotes the matrix resulting from the application of convolutional layers to the image training set $\Xmat$. 

\section{Variational Objective for the Proposed \cnngprf Model} 

Consider the proposed \cnngprf model. 
In the main paper we report an expression for the lower bound on the marginal likelihood when treating $\Wmat$ and $\convweights$ variationally, while assuming $\Omegamat$ sampled from the prior (which in turn depends on covariance parameters $\thetavect$).
Assume now that we are interested in carrying out inference over $\Omegamat$, as well as $\Wmat$ and $\convweights$.
Extending the expression of the variational lower bound in the main paper, we can introduce an approximate posterior over $\Omegamat$, say $q(\Omegamat)$ and attempt to optimize it within the \mcd framework. 
The lower bound to the log-marginal likelihood $\LL_{\Wmat,\convweights,\Omegamat} = \log \left[p(\Ymat | \Xmat, \thetavect\right]$ can be expressed as
\begin{equation}
\label{eq:elbo:varOmega}
\LL_{\Wmat,\convweights,\Omegamat} \geq 
\E_{q(\Wmat, \convweights, \Omegamat)} \left( \log\left[ p\left(\Ymat | \Xmat, \Wmat, \convweights, \Omegamat, \thetavect\right) \right] \right)
- \mathrm{KL}\left[q\left(\Wmat, \convweights, \Omegamat\right) \| p\left(\Wmat, \convweights, \Omegamat | \thetavect\right) \right]
\end{equation}
We can again apply \mcd by introducing Bernoulli variables reparameterizing 
\begin{equation}
\Wmat = \Mmat_w \, \diag[\zvect_w] \qquad \mathrm{with} \qquad (\zvect_w)_i \sim \mathrm{Bernoulli}(\pi_w) \text{,}
\end{equation}
and similarly for $\convweights$ and $\Omegamat$.

Again, the expectation can be unbiasedly estimated using Monte Carlo and also considering a mini-batch of size $m$:
\begin{equation}
\E_{q(\Wmat, \convweights, \Omegamat)} \left( \log\left[ p\left(\Ymat | \Xmat, \Wmat, \convweights, \Omegamat, \thetavect\right) \right] \right)
\approx
\frac{\nobs}{m} \frac{1}{N_{\mathrm{MC}}} \sum_{i = 1}^{N_{\mathrm{MC}}} \sum_{k \in \mathcal{I}_{m}} \log\left[ p\left(\yvect_k | \xvect_k, \Wmat^{(i)}, \convweights^{(i)}, \Omegamat^{(i)}, \thetavect\right) \right] \text{,}
\end{equation}
with $\Wmat^{(i)}, \convweights^{(i)}, \Omegamat^{(i)} \sim q(\Wmat, \convweights, \Omegamat)$, and $\mathcal{I}_m$ is a set of $m$ indices to select a mini-batch of training points.

The KL term can be approximated following \cite{Gal16}, noting that the fact that we are treating $\Omegamat$ variationally, gives rise to extra terms that involve the \gp length-scale $\ell$:
\begin{equation}
\mathrm{KL}\left[q\left(\Wmat, \convweights, \Omegamat\right) \| p\left(\Wmat, \convweights, \Omegamat | \thetavect\right) \right]
\approx
\frac{\pi_w}{2} \left\| \Mmat_w \right\|^2 + 
\frac{\pi_{\psi}}{2} \left\| \Mmat_\psi \right\|^2 + 
\frac{\ell^2 \pi_{\Omega}}{2} \left\| \Mmat_\Omega \right\|^2 + 
N_{\mathrm{RF}} \, \nconv \, \log\left( \ell^{-2} \right)
\end{equation}

When using this expression to optimize all variational parameters pertaining to $q\left(\Wmat, \convweights, \Omegamat\right)$ jointly with $\thetavect$ we encountered some instabilities, and therefore we decided to report results when fixing the covariance parameters $\thetavect$.

For the case where $\Omegamat$ is not learned variationally, which is what we do when employing the \sorf approximation, we can simply draw $\Omegamat$ from the prior $\norm(\Omegamat_{\cdot j} | \zerovect, \Lambda^{-1})$ and consider the reparameterization \cite{Gredilla10}:
\begin{equation}
\Omegamat_{\cdot j} = \Lambda^{-\frac{1}{2}} \varepsilonvect \text{,}
\end{equation}
where $\varepsilon_i \sim \norm(\varepsilon_i | 0, 1)$.
This reparameterization allows for the update of covariance parameters $\thetavect$ fixing the randomness in the sampling from $p(\Omegamat | \thetavect)$.
The results comparing \cnngpsorf when updating or fixing $\thetavect$ throughout optimization are reported in table~\ref{tab:learn:theta:sorf}.
It is interesting to notice how fixing covariance parameters $\thetavect$ leads to comparable performance to the case where we learn them.


\begin{table}[t]
\small
  \caption{\label{tab:learn:theta:sorf}
Results on the proposed \cnngpsorf when fixing or learning covariance parameters $\thetavect$. 
All results were obtained on \mnist, \cifart, and \cifarh without subsampling the data. 
Please refer to table~1 in the main paper for details on the convolutional structure corresponding to ``SHALLOW'' and ``DEEP''.}
\centering
\begin{tabular}{p{1cm} |p{1cm} p{1cm} | p{1cm} p{1cm} | p{1cm} p{1cm} | p{1cm} p{1cm} }
\toprule
\multicolumn{1}{c|}{} & \multicolumn{4}{c|}{SHALLOW} & \multicolumn{4}{c}{DEEP}\\ 
\midrule
\multicolumn{1}{c|}{} & \multicolumn{2}{c|}{\mnist} & \multicolumn{2}{c|}{\cifart} & \multicolumn{2}{c|}{\cifart} & \multicolumn{2}{c}{\cifarh}\\
\midrule
Metrics & Fixed & Learned & Fixed & Learned & Fixed & Learned & Fixed & Learned \\
\midrule
\err & 0.006 & 0.005 & 0.203 & 0.192 & 0.113 & 0.115 & 0.352 & 0.359\\
\mnll & 0.018 & 0.018 & 0.610 & 0.584 & 0.348 & 0.355 & 1.264 & 1.287\\
\ece & 0.002 & 0.003 & 0.015 & 0.010 & 0.051 & 0.054 & 0.050 & 0.054\\
\brier & 0.009 & 0.008 & 0.288 & 0.271 & 0.170 & 0.173 & 0.466 & 0.478\\
\bottomrule
\end{tabular}
\end{table}

\section{Variational inference of filters in \gpdnn}

In this section we report results when applying variational inference on the weights in \gpdnn \cite{Bradshaw17}.
In order to do this, we implemented \mcd for the convolutional parameters, similarly to what presented in the main paper for our \cnngprf model. 

\begin{table}[h] 
\small
  \caption{\label{tab:gpdnn:mcd:conv}
Results on the proposed \cnngpsorf vs \gpdnn when inferring convolutional parameters using \mcd. 
All results were obtained on \mnist, \cifart, and \cifarh without subsampling the data. 
Please refer to table~1 in the main paper for details on the convolutional structure corresponding to ``SHALLOW'' and ``DEEP''.}
\centering
\begin{tabular}{ p{0.8cm}|p{1.4cm} p{0.8cm}|p{1.4cm} p{0.8cm}|p{1.4cm} p{0.8cm}|p{1.4cm} p{0.8cm}  }
\toprule
\multicolumn{1}{c|}{} & \multicolumn{4}{c|}{SHALLOW} & \multicolumn{4}{c}{DEEP}\\ 
\midrule
\multicolumn{1}{c|}{} & \multicolumn{2}{c|}{\mnist} & \multicolumn{2}{c|}{\cifart} & \multicolumn{2}{c|}{\cifart} & \multicolumn{2}{c}{\cifarh}\\
\midrule
Metrics & \cnngprf & \gpdnn & \cnngprf & \gpdnn & \cnngprf & \gpdnn & \cnngprf & \gpdnn \\
\midrule
\err & 0.005 & 0.005 & 0.172 & 0.172 & 0.111 & 0.190 & 0.351 & 0.820\\
\mnll & 0.014 & 0.019 & 0.535 & 0.531 & 0.344 & 0.675 & 1.255 & 8.606\\
\ece & 0.004 & 0.005 & 0.012 & 0.012 & 0.051 & 0.036 & 0.050 & 0.527\\
\brier & 0.0071 & 0.008 & 0.245 & 0.244 & 0.168 & 0.278 & 0.466 & 1.268\\
\bottomrule
\end{tabular}
\end{table}

The results in table~\ref{tab:gpdnn:mcd:conv} indicate that this improves the calibration and accuracy of \gpdnn compared to optimizing the filters. 
In the case of a shallow convolutional architecture, the performance of \cnngprf and \gpdnn are comparable, although in the deeper case \cnngprf achieves better performance.
This supports the intuition that inferring convolutional parameters, ranther than optimizing them, leads to considerable improvements in calibration.

\section{Reliability diagrams for \cnngprf and \name{cnn+mcd}}

In figure~\ref{reliability diagrams} we report the reliability diagram of our \cnngprf model and \name{cnn+mcd}.
These plots show that our approach and Bayesian \cnns are well-calibrated.

\begin{figure}[h]
\vskip 0.2in
\begin{center}
\begin{tikzpicture}

\node[inner sep=0pt](mnist-lenet-mnll) at (2.5, 5){\includegraphics[width=3.2cm]{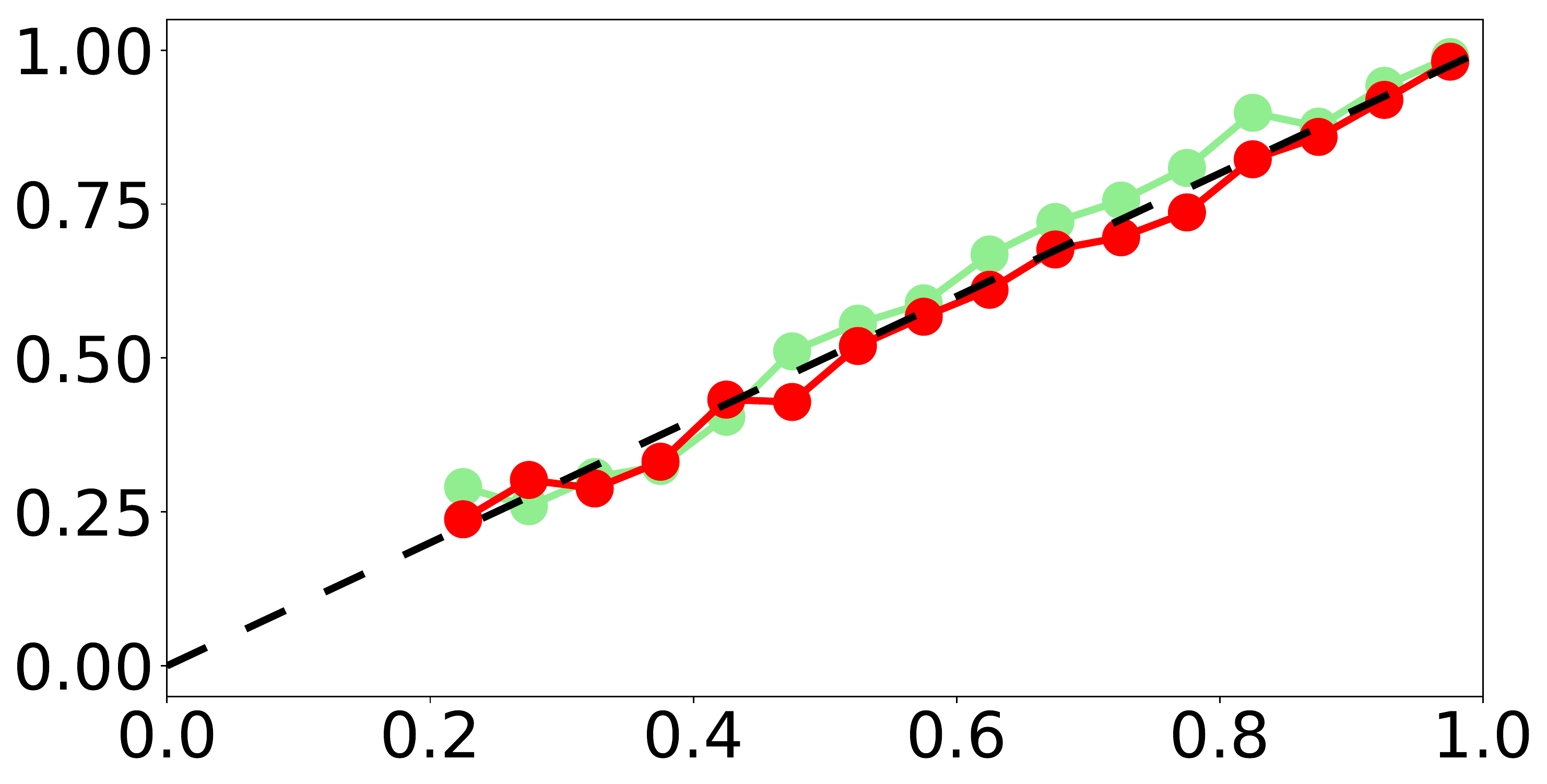}};

\node[inner sep=0pt](mnist-lenet-ece) at (7.5, 5){\includegraphics[width=3.2cm]{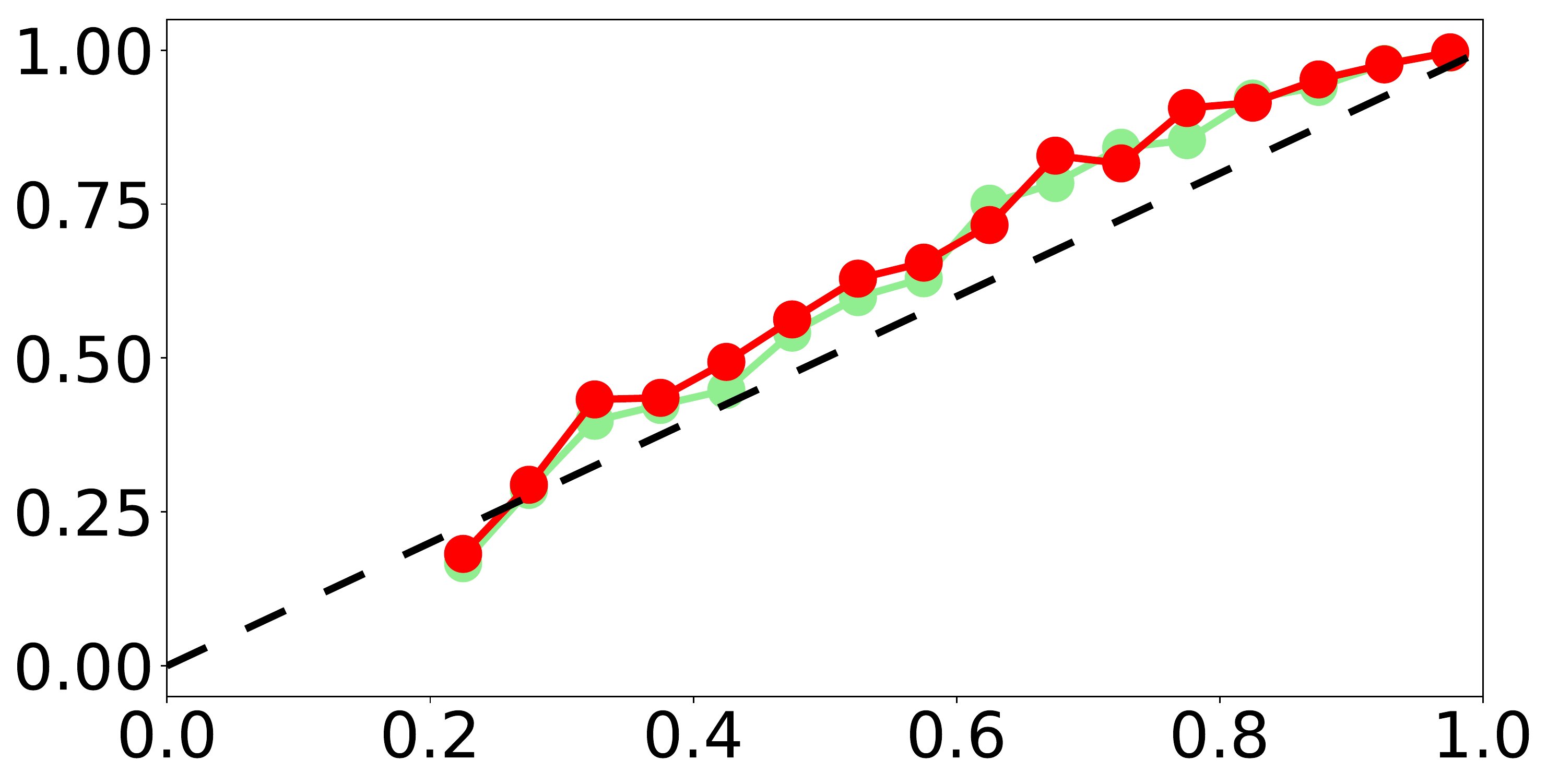}};

\node[inner sep=0pt](mnist-lenet-ece) at (12.5, 5){\includegraphics[width=3.2cm]{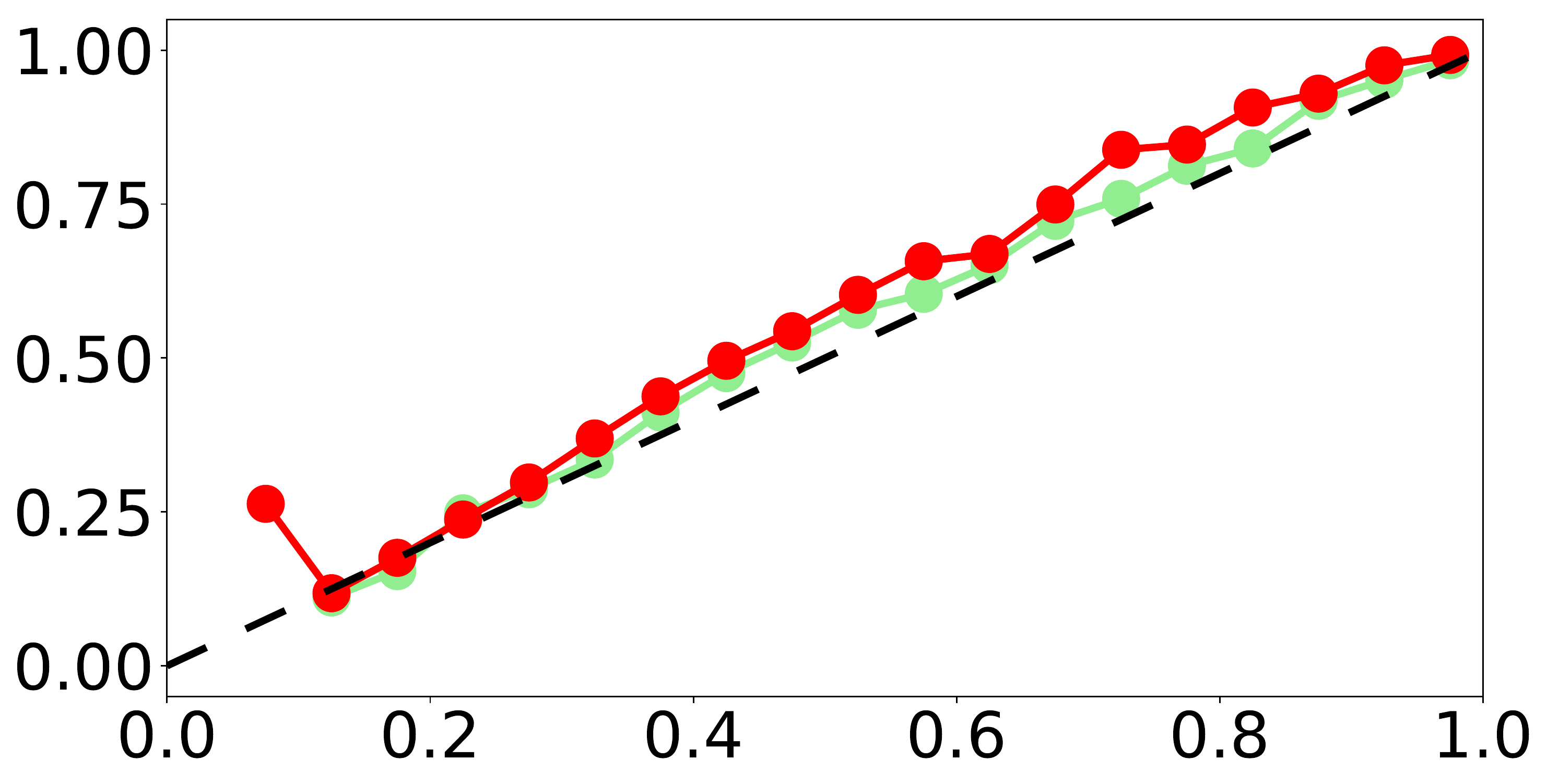}};

\node(text) at (2.5, 4.0) {\scriptsize $\mathrm{confidence}$};
\node(text) at (7.5, 4.0) {\scriptsize $\mathrm{confidence}$};
\node(text) at (12.5, 4.0) {\scriptsize $\mathrm{confidence}$};

\node[rotate=90](text) at (0.7, 5) {\scriptsize $\mathrm{accuracy}$};
\node[rotate=90](text) at (5.7, 5) {\scriptsize $\mathrm{accuracy}$};
\node[rotate=90](text) at (10.7,5) {\scriptsize $\mathrm{accuracy}$};




\end{tikzpicture}
\caption{Reliability diagrams for \cnngprf (red) and \name{cnn+mcd} \cite{Gal16b} (green) 
Left: \cifart - \lenet
Center: \cifart - shallow \resnet.
Right: \cifarh - deep \resnet. 
See table~1 in the main paper for details on the shallow and deep convolutional architectures that we use in this experiments.
}
\label{reliability diagrams}
\end{center}
\vskip -0.2in
\end{figure}
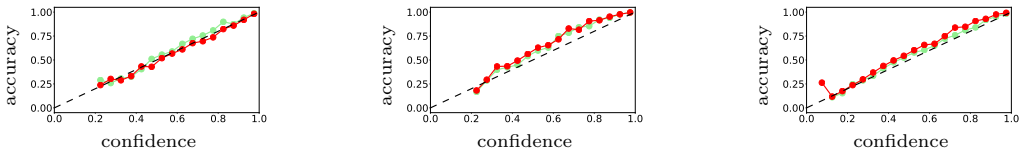

\end{document}